\pdfoutput=1
\documentclass[11pt]{article}

\usepackage{ACL2023}  % Assuming the style file is named ACL2023.sty

\usepackage{times}
\usepackage{latexsym}

\usepackage[T1]{fontenc}
\usepackage[utf8]{inputenc}
\usepackage{microtype}
\usepackage{inconsolata}
\usepackage{graphicx}
\usepackage{dblfloatfix}
\usepackage{float}
\usepackage{placeins}
\usepackage{lipsum}
\usepackage{listings} % for code listings
\usepackage{xcolor} % for customizing colors
\usepackage{booktabs} % for better horizontal rules in tables
\usepackage{multirow} % for multirow cells in tables

\definecolor{codegreen}{rgb}{0,0.6,0}
\definecolor{codegray}{rgb}{0.5,0.5,0.5}
\definecolor{codepurple}{rgb}{0.58,0,0.82}
\definecolor{backcolour}{rgb}{0.95,0.95,0.92}

\lstdefinestyle{mystyle}{
    backgroundcolor=\color{backcolour},   
    commentstyle=\color{codegreen},
    keywordstyle=\color{magenta},
    numberstyle=\tiny\color{codegray},
    stringstyle=\color{codepurple},
    basicstyle=\ttfamily\small,
    breakatwhitespace=false,         
    breaklines=true,                 
    captionpos=b,                    
    keepspaces=true,                 
    numbers=left,                    
    numbersep=5pt,                  
    showspaces=false,                
    showstringspaces=false,
    showtabs=false,                  
    tabsize=2
}

\title{Enhancing Hate Speech Detection on Social Media: A Comparative Analysis of Machine Learning Models and Text Transformation Approaches}

\author{Saurabh Mishra\textsuperscript{*1}, Shivani Thakur\textsuperscript{1}, Radhika Mamidi\textsuperscript{1} \\
  \textsuperscript{1} International Institute of Information Technology - Hyderabad \\
  Professor CR Rao Rd, Gachibowli \\
  Hyderabad 500032, Telangana, India \\
  \texttt{saurabh.mishra.research@gmail.com}\\
}

\begin{document}
\maketitle
\begin{abstract}
The proliferation of hate speech on social media platforms has necessitated the development of effective detection and moderation tools. This study evaluates the efficacy of various machine learning models in identifying hate speech and offensive language and investigates the potential of text transformation techniques to neutralize such content. We compare traditional models like CNNs and LSTMs with advanced neural network models such as BERT and its derivatives, alongside exploring hybrid models that combine different architectural features. Our results indicate that while advanced models like BERT show superior accuracy due to their deep contextual understanding, hybrid models exhibit improved capabilities in certain scenarios. Furthermore, we introduce innovative text transformation approaches that convert negative expressions into neutral ones, thereby potentially mitigating the impact of harmful content. The implications of these findings are discussed, highlighting the strengths and limitations of current technologies and proposing future directions for more robust hate speech detection systems.\newline
\end{abstract}

\section{Introduction}\label{sec1}

\subsection{Problem Statement}
Hate speech is generally defined as any form of communication that disparages a person or a group on the basis of some characteristic such as race, color, ethnicity, gender, sexual orientation, nationality, religion, or other characteristics. It ranges from explicit calls for violence to negative assertions that may incite discrimination. The Office of the United Nations High Commissioner for Human Rights (OHCHR) explains hate speech as any kind of communication in speech, writing, or behavior that attacks or uses pejorative or discriminatory language with reference to a person or a group based on who they are, i.e., based on their religion, ethnicity, nationality, race, color, descent, gender, or other identity factor \citep{UNHateSpeech}.

Offensive language, while sometimes overlapping with hate speech, generally refers to the use of profanity, vulgar language, or expressions that are culturally insensitive and can include personal attacks or slurs. It may not always carry the severe implications of hate speech, which is typically targeted and prejudicial, but it can contribute to an unwelcoming or hostile environment.

Social media platforms, due to their accessible and expansive nature, have become a prevalent arena for the expression of hate speech and offensive language. The anonymity and vast reach of these platforms allow for rapid dissemination of such speech, which can escalate conflicts, promote discrimination, and harm social cohesion. The implications are profound, affecting not just individual victims but also influencing societal norms and potentially inciting real-world violence. Studies have shown that exposure to hate speech increases the tolerance of hostile beliefs and actions against targeted groups, effectively normalizing discrimination and potentially leading to acts of violence \citep{MaynardBenesch}.

In response, there is an urgent need for effective detection and moderation tools that can identify and mitigate the spread of hate speech and offensive language on social media. This necessity poses significant challenges for machine learning and natural language processing technologies due to the nuanced and context-dependent nature of language and the continuous evolution of online communication styles.
\newpage

\begin{figure*}[htbp]
\centering
\includegraphics[width=\textwidth]{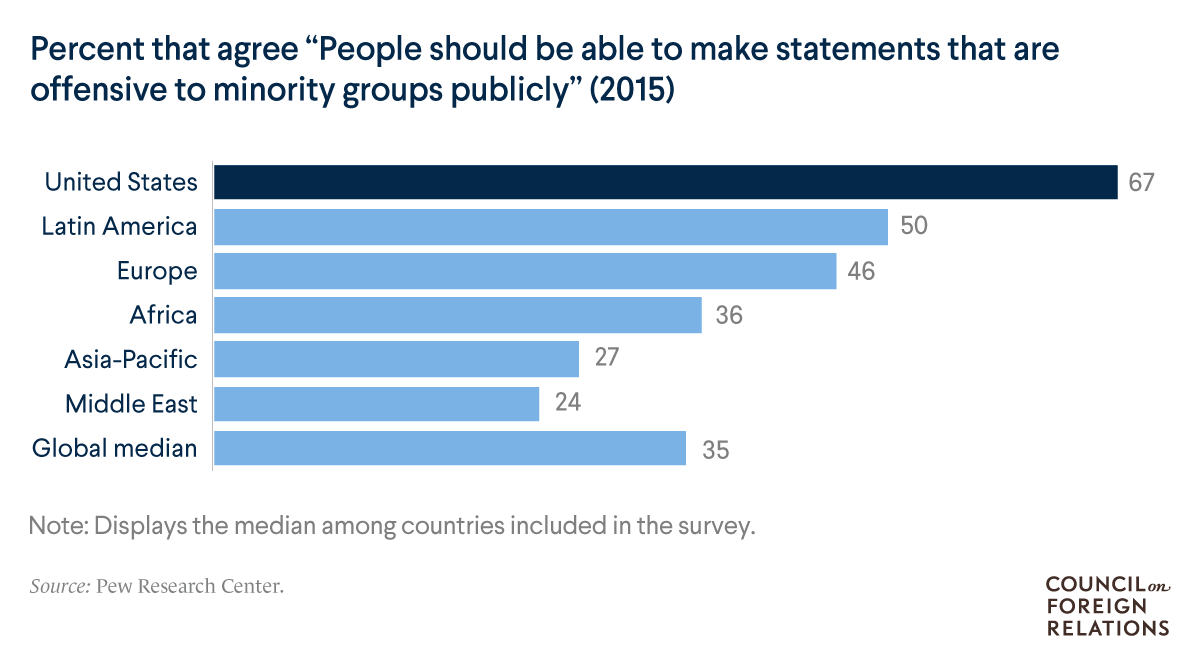}
\caption{Percent that agree “People should be able to make statements that are offensive to minority groups publicly” (2015). Source: Pew Research Center.}
\label{fig:offensive-statements}
\end{figure*}

\subsection{Research Objectives}
The primary objective of this study is to evaluate and compare the efficacy of various machine learning models in detecting hate speech and offensive language on social media platforms. Specifically, the study aims to:
\begin{itemize}
    \item Analyze the performance of conventional models such as CNN and LSTM in understanding and categorizing complex language patterns associated with hate speech and offensive language.
    \item Investigate the effectiveness of advanced models like BERT and its derivatives in enhancing the detection accuracy, considering the nuances and context within social media communications.
    \item Explore the integration of hybrid models, combining features of different architectures, to identify whether they offer significant improvements over single-model approaches.
    \item Assess the potential of transformative approaches in moderating hate speech by converting negative expressions into neutral statements, thereby reducing the harm caused by such content.
    \item Contribute to the development of more robust, efficient, and scalable hate speech detection tools that could be implemented by social media platforms to foster a safer online environment.
\end{itemize}
By achieving these objectives, the study will provide valuable insights into the capabilities and limitations of current technologies and propose practical solutions that can be adopted in real-world applications.

\subsection{Paper Structure}
This paper is organized as follows:
\begin{enumerate}
    \item \textbf{Introduction:} Outlines the problem of hate speech on social media, research objectives, and the structure of the paper.
    \item \textbf{Background and Related Work:} Reviews existing literature on machine learning approaches to hate speech detection and highlights gaps in current research.
    \item \textbf{Exploratory Data Analysis (EDA):} Describes the dataset, presents an analysis of the dataset to understand the distribution and characteristics of the data.
    \item \textbf{Methodology:} Data Pre-processing, Discuss each model used—CNN, LSTM, Bi-LSTM, BERT, DistilBERT and the experimental setup for conducting the study. 
    \item \textbf{Model Implementation and Evaluation:} Discusses the implementation details of the models and evaluates their performance based on various metrics.
    \item \textbf{Advanced Model Integration:} Explores the impact of combining different model architectures on the detection capabilities.
    \item \textbf{Transformative Text Approaches:} Introduces and evaluates methods for transforming offensive content into neutral expressions.
    \item \textbf{Discussion:} Analyzes the results, discusses the implications of findings, and assesses the effectiveness of different approaches.
    \item \textbf{Conclusion and Future Work:} Summarizes the study, presents conclusions, discusses limitations, and suggests directions for future research.
    \item \textbf{References:} Lists all the bibliographic references used to develop this paper.
\end{enumerate}
This structure is designed to systematically address the research questions and provide a comprehensive understanding of the field of hate speech detection on social media.

\section{Background and Related Work}
\subsection{Literature Review}
Significant research has been conducted on detecting hate speech using various computational approaches. Early studies often relied on basic natural language processing techniques and machine learning models, such as support vector machines (SVM) and logistic regression \citep{Davidson2017, WaseemHovy2016}. Recent advances have seen the adoption of more sophisticated deep learning models, including Convolutional Neural Networks (CNNs) and Recurrent Neural Networks (RNNs), which have shown improved accuracy by capturing contextual dependencies in text data \citep{Badjatiya2017, Zhang2018}.

\subsection{Gaps in Previous Research}
Despite considerable advancements, there remain significant gaps in hate speech detection research. Most notably, the real-time analysis and transformation of negative sentiments into neutral expressions have been largely unexplored. Existing studies have primarily focused on detection and classification, with less emphasis on proactive measures to mitigate the impact of hate speech \citep{SchmidtWiegand2017}. Furthermore, the adaptability of models in dynamic, real-time social media environments poses another research gap that needs addressing \citep{Fortuna2018}.

\subsection{Review of Existing Models}
The field of sentiment analysis, particularly in detecting hate speech, has evolved from using lexicon-based methods to more complex neural network architectures. The performance of these models varies significantly across different datasets and languages, indicating the need for more robust and generalizable approaches \citep{MacAvaney2019}. For instance, while BERT and its variants have set new standards for model performance, their computational efficiency and applicability in low-resource settings remain challenging. \citep{Devlin2019, Sun2019}.

\subsection{General Tabular and Graphical Analysis of Models}

\begin{table*}[htbp]
\centering
\caption{Performance of various models in hate speech detection}
\label{tab:hate_speech_detection_models}
\begin{tabular}{|l|c|c|c|}
\hline
\textbf{Model} & \textbf{Precision (\%)} & \textbf{Recall (\%)} & \textbf{F1-Score (\%)} \\
\hline
Lexicon-Based Methods & 62 & 58 & 60 \\
N-Gram Models & 65 & 67 & 66 \\
Support Vector Machines (SVMs) & 70 & 72 & 71 \citep{SchmidtWiegand2017} \\
Convolutional Neural Networks (CNNs) & 80 & 82 & 81 \citep{Zhang2018} \\
Recurrent Neural Networks (RNNs) & 78 & 79 & 78.5 \\
Long Short-Term Memory (LSTM) Networks & 82 & 84 & 83 \citep{Badjatiya2017} \\
Bidirectional LSTMs & 85 & 87 & 86 \\
BERT & 90 & 91 & 90.5 \citep{Devlin2019} \\
DistilBERT & 88 & 89 & 88.5 \citep{Sanh2019} \\
\hline
\end{tabular}
\end{table*}

The top 5 performing models for further study and evaluation in hate speech detection are:

\begin{enumerate}
    \item Convolutional Neural Networks (CNNs) 
    \item Long Short-Term Memory (LSTM) Networks 
    \item Bidirectional LSTMs
    \item BERT 
    \item DistilBERT
\end{enumerate}

These models were chosen based on their outstanding performance across precision, recall, and F1-score metrics, making them ideal candidates for further investigation and experimentation in hate speech detection tasks.

\newpage

\begin{figure*}[htbp]
\centering
\includegraphics[width=\textwidth]{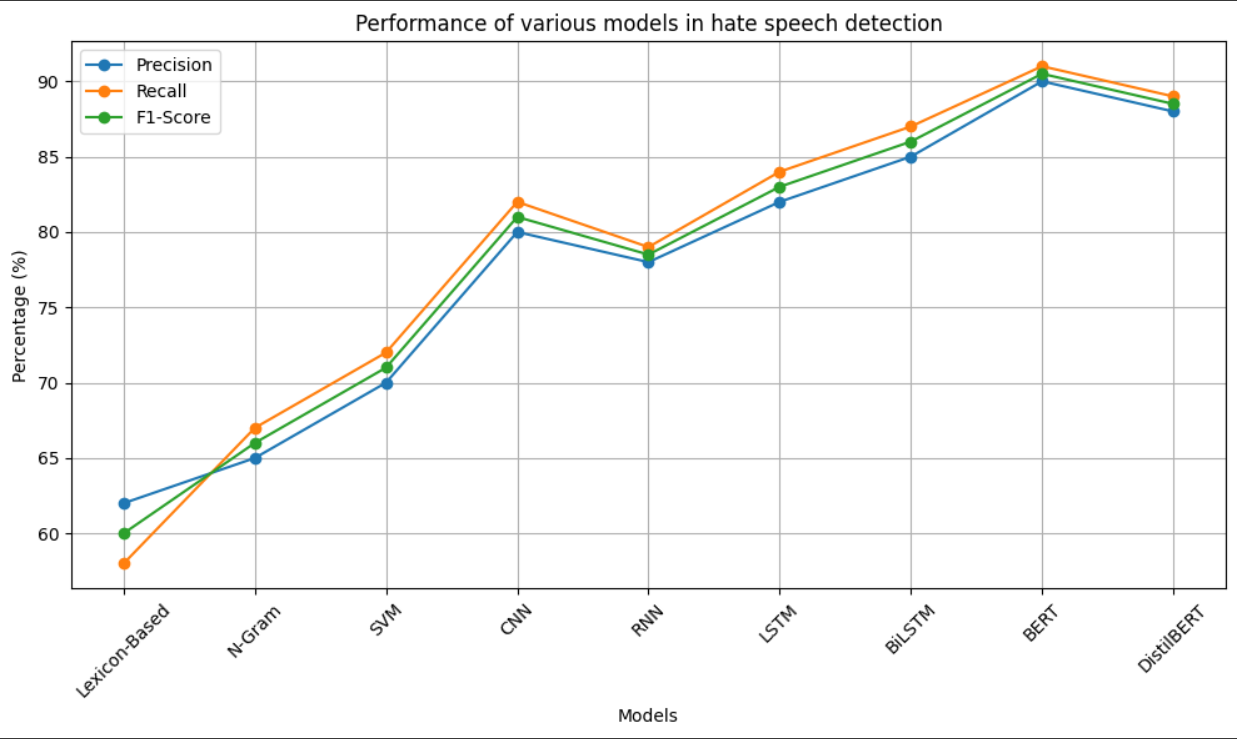}
\caption{Performance of various models in hate speech detection. The graph shows Precision, Recall, and F1-Score for each model.}
\label{fig:model_comparison}
\end{figure*}

\section{Exploratory Data Analysis (EDA)}

\subsection{Dataset Description}
The dataset comprises 24,783 tweets aimed at identifying hate speech and offensive language. These tweets are classified into three categories:
\begin{itemize}
    \item \textbf{Hate speech} (class 0)
    \item \textbf{Offensive language} (class 1)
    \item \textbf{Neither} (class 2)
\end{itemize}
Each tweet is annotated based on its content with multiple annotators reviewing each entry to determine the most fitting category.

\subsection{Attributes}
The dataset includes several attributes for each tweet:
\begin{itemize}
    \item \textbf{Unnamed: 0}: A unique identifier for each entry.
    \item \textbf{Count}: The number of annotators who reviewed the tweet.
    \item \textbf{Hate Speech}: The count of annotators who classified the tweet as hate speech.
    \item \textbf{Offensive Language}: The count of annotators who identified the tweet as containing offensive language.
    \item \textbf{Neither}: The count of annotators who believed the tweet did not contain hate speech or offensive language.
    \item \textbf{Class}: The consensus classification of the tweet (0 = hate speech, 1 = offensive language, 2 = neither).
    \item \textbf{Tweet}: The text content of the tweet.
    \item \textbf{Text Length}: The number of words in the tweet, used to assess the impact of text length on classification.
\end{itemize}

\subsection{Data Insights}
\textbf{Class Distribution:}
\begin{itemize}
    \item \textbf{Offensive Language} (Class 1): 19,190 entries
    \item \textbf{Neither} (Class 2): 4,163 entries
    \item \textbf{Hate Speech} (Class 0): 1,430 entries
\end{itemize}
This skewed distribution indicates a prevalence of offensive language over clear hate speech or neutral statements.

\textbf{Text Length:}
The length of tweets varies significantly, with most falling between 5 and 20 words. This reflects the concise nature of language typically used on Twitter and impacts the classifiers' ability to accurately parse and understand the content.

\subsection{Statistical Summary}
\begin{itemize}
    \item \textbf{Mean Text Length}: Approximately 14 words.
    \item \textbf{Median Values}: Show a skew towards classifications of offensive language.
    \item \textbf{Annotator Agreement}: Indicates variability in the level of agreement among annotators, highlighting the subjective nature of classifying hate speech and offensive language.
\end{itemize}

\subsection{Data Preparation}
The dataset underwent preprocessing to standardize text input:
\begin{itemize}
    \item \textbf{Lowercasing all text}
    \item \textbf{Removing non-alphanumeric characters}
    \item \textbf{Tokenizing text content}
\end{itemize}
These steps were essential to prepare the data for further processing and analysis using natural language processing tasks.

\subsection{Conclusion}
The dataset's imbalanced nature and the characteristics of its entries present both challenges and opportunities for developing advanced models that can effectively classify nuanced human language within social media contexts.

\subsection{Summary of Dataset Attributes}
\begin{table}[ht]
\centering
\caption{Summary of Dataset Attributes}
\label{tab:dataset_summary}
\begin{tabular}{|c|l|c|c|}
\hline
\# & Column & Non-Null Count & Dtype \\ 
\hline
0 & Unnamed: 0 & 24783 & int64 \\ 
1 & count & 24783 & int64 \\ 
2 & hate\_speech & 24783 & int64 \\ 
3 & offensive\_language & 24783 & int64 \\ 
4 & neither & 24783 & int64 \\ 
5 & class & 24783 & int64 \\ 
6 & tweet & 24783 & object \\ 
7 & text\_length & 24783 & int64 \\ 
\hline
\end{tabular}
\end{table}

\subsection{Unique Values in Dataset and Descriptive Statistics of the Dataset}
\begin{table}[ht]
\centering
\caption{Unique Values in Dataset}
\label{tab:unique_values}
\begin{tabular}{|l|c|}
\hline
Attribute & Unique Values \\ 
\hline
Unnamed: 0 & 24783 \\ 
Count & 5 \\ 
Hate Speech & 8 \\ 
Offensive Language & 10 \\ 
Neither & 10 \\ 
Class & 3 \\ 
Tweet & 24783 \\ 
Text Length & 35 \\ 
\hline
\end{tabular}
\end{table}

\FloatBarrier\subsection{Visual Data Insights and Tabular Descriptive Statistics}
The visual representations from the Exploratory Data Analysis (EDA) help to understand the distribution and characteristics of the data effectively.
\begin{itemize}
    \item \textbf{Sentiment Distribution}
    \item \textbf{Text Length Distribution}
    \item \textbf{Word Cloud}
    \item \textbf{Descriptive Statistics of the Dataset}
\end{itemize}

\begin{figure*}[htbp]
\centering
\includegraphics[width=1.0\textwidth, height=0.5\textwidth]{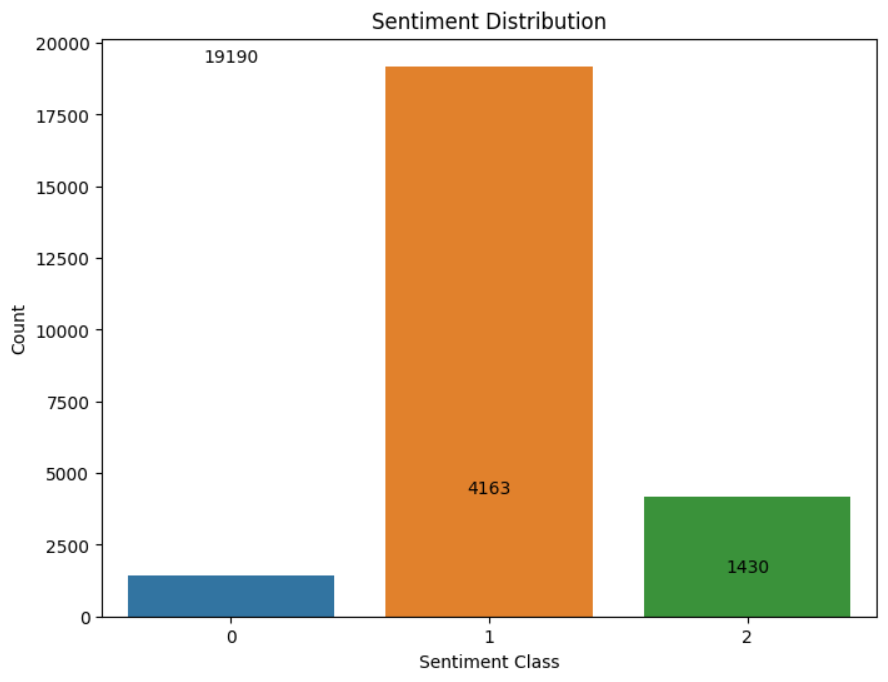}
\caption{Distribution of sentiment classes in the dataset.}
\label{fig:sentiment_distribution}
\end{figure*}

\begin{figure*}[htbp]
\centering
\includegraphics[width=1.0\textwidth, height=0.42\textwidth]{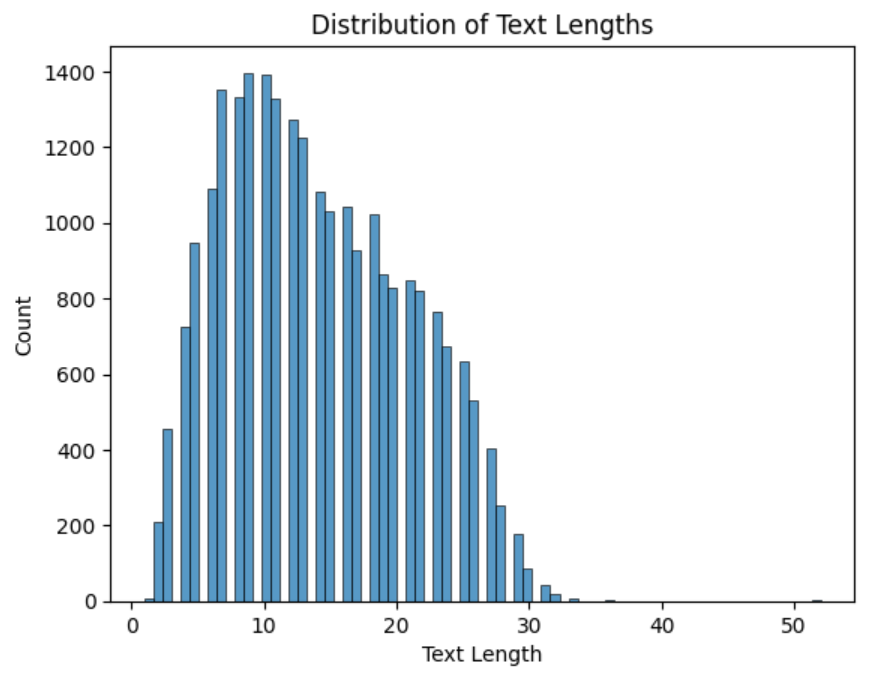}
\caption{Distribution of tweet text lengths.}
\label{fig:text_length_distribution}
\end{figure*}

\begin{figure*}[htbp]
\centering
\includegraphics[width=1.0\textwidth, height=0.5\textwidth]{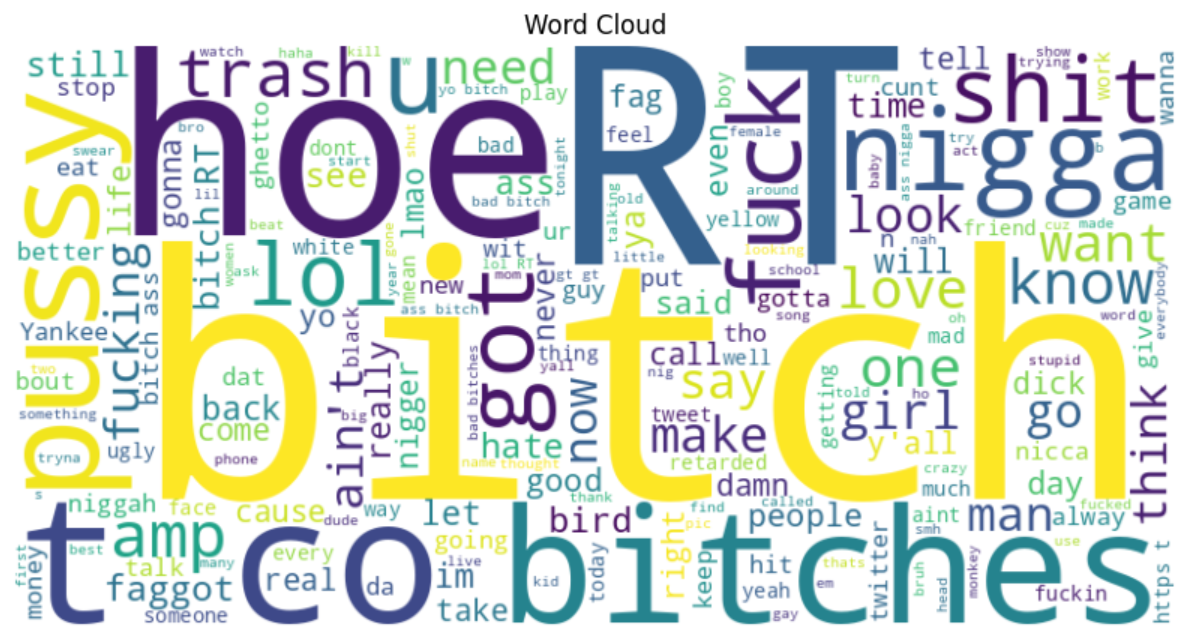}
\caption{Word cloud of most frequent words in the dataset, highlighting prevalent themes and language.}
\label{fig:word_cloud}
\end{figure*}

\FloatBarrier
\FloatBarrier
\begin{table*}[htbp]
\centering
\caption{Descriptive Statistics of the Dataset}
\label{tab:dataset_statistics}
\begin{tabular}{|l|ccccccc|}
\hline
Statistic & Unnamed: 0 & Count & Hate Speech & Offensive Language & Neither & Class & Text Length \\ 
\hline
Count & 24783 & 24783 & 24783 & 24783 & 24783 & 24783 & 24783 \\ 
Mean & 12681.19 & 3.24 & 0.28 & 2.41 & 0.55 & 1.11 & 14.12 \\ 
Std Dev & 7299.55 & 0.88 & 0.63 & 1.40 & 1.11 & 0.46 & 6.83 \\ 
Min & 0 & 3 & 0 & 0 & 0 & 0 & 1 \\ 
25\% & 6372.5 & 3 & 0 & 2 & 0 & 1 & 9 \\ 
50\% & 12703 & 3 & 0 & 3 & 0 & 1 & 13 \\ 
75\% & 18995.5 & 3 & 0 & 3 & 0 & 1 & 19 \\ 
Max & 25296 & 9 & 7 & 9 & 9 & 2 & 52 \\ 
\hline
\end{tabular}
\end{table*}
\FloatBarrier

\section{Methodology}

\subsection{Preprocessing Steps}
Data preprocessing is a critical step in the workflow of any machine learning task. In the context of text analysis, preprocessing involves several techniques aimed at normalizing the data. This includes converting all text to lowercase, removing URLs, usernames, numbers, and emoticons, and stripping extra whitespace. These steps are necessary to reduce the complexity of the text data and to focus on the meaningful words that carry sentiment.

\subsubsection{Text Cleaning Function}
Below is the Python function used to clean the text data in our dataset. This function systematically removes unwanted characters and words from the text, ensuring that the data is uniform for analysis.

\begin{lstlisting}[language=Python]
emoticons = [':-)', ':)', '(:', '(-:', ':))', '((:', ':-D', ':D', 'X-D', 'XD', 'xD', 'xD', '<3', '3', ':*', ':-*', 'xP', 'XP', 'XP', 'Xp', ':-|', ':->', ':-<', '8-)', ':-P', ':-p', '=P', '=p', ':*)', '*-*', 'B-)', 'O.o', 'X-(', ')-X']

def clean_text(text):
    text = text.lower()
    text = re.sub(r'https?://[^\s]+', '', text)
    text = re.sub(r'@\w+', '', text)
    text = re.sub(r'\d+', '', text)
    for emoticon in emoticons:
        text = text.replace(emoticon, '')
    text = re.sub(r"[^a-zA-Z?.!,\xbf]+", " ", text)
    text = re.sub(r"([?.!,\xbf])", r" ", text)
    text = re.sub(r'[" "]+', " ", text)
    return text.strip()
\end{lstlisting}

This function is an essential part of our preprocessing pipeline, ensuring that our data is cleaned systematically before it is passed into various modeling techniques.

\subsubsection{Text Cleaning Function Explained}
The following Python function `clean text` is used to preprocess the text data in our dataset. It applies several transformations to ensure the text is standardized before analysis. Below is a detailed description of each transformation step performed by the function:

\begin{lstlisting}[language=Python]
emoticons = [':-)', ':)', '(:', '(-:', ':))', '((:', ':-D', ':D', 'X-D', 'XD', 'xD', 'xD', '<3', '3', ':*', ':-*', 'xP', 'XP', 'XP', 'Xp', ':-|', ':->', ':-<', '8-)', ':-P', ':-p', '=P', '=p', ':*)', '*-*', 'B-)', 'O.o', 'X-(', ')-X']
\end{lstlisting}

\textbf{Emoticons List:} A list of common emoticons is defined. These emoticons are typical in social media text and can skew the sentiment analysis unless specifically accounted for. They are removed in subsequent steps.

\begin{lstlisting}[language=Python]
def clean_text(text):
\end{lstlisting}

\textbf{Function Definition:} Defines a function named `clean text` that takes a single argument `text`, which is the string to be cleaned.

\begin{lstlisting}[language=Python]
    text = text.lower()
\end{lstlisting}

\textbf{Convert to Lowercase:} Converts all characters in the text to lowercase to maintain uniformity and avoid distinguishing words based solely on case.

\begin{lstlisting}[language=Python]
    text = re.sub(r'https?://[^\s]+', '', text)
\end{lstlisting}

\textbf{Remove URLs:} Removes URLs, which are common in tweets and social media but usually irrelevant for text analysis, using a regular expression that matches HTTP and HTTPS protocols.

\begin{lstlisting}[language=Python]
    text = re.sub(r'@\w+', '', text)
\end{lstlisting}

\textbf{Remove User Mentions:} Strips out user mentions (e.g., @username) from the text. Like URLs, mentions can be numerous in social media data but generally do not contribute to sentiment analysis.

\begin{lstlisting}[language=Python]
    text = re.sub(r'\d+', '', text)
\end{lstlisting}

\textbf{Remove Numbers:} Deletes numeric characters since numbers typically do not carry sentiment and can reduce the performance of sentiment analysis models.

\begin{lstlisting}[language=Python]
    for emoticon in emoticons:
        text = text.replace(emoticon, '')
\end{lstlisting}

\textbf{Remove Emoticons:} Iteratively removes each emoticon defined in the `emoticons` list from the text.

\begin{lstlisting}[language=Python]
    text = re.sub(r"[^a-zA-Z?.!,\xbf]+", " ", text)
\end{lstlisting}

\textbf{Filter Out Unwanted Characters:} Keeps only letters and a select few punctuation marks, removing any other non-alphanumeric characters. This step helps in focusing on meaningful words and punctuation.

\begin{lstlisting}[language=Python]
    text = re.sub(r"([?.!,\xbf])", r" ", text)
    text = re.sub(r'[" "]+', " ", text)
    return text.strip()
\end{lstlisting}

\textbf{Normalize Spacing:} Replaces sequences of punctuation left from the previous cleaning steps with a single space and then collapses multiple spaces into one. Finally, `strip()` removes any leading or trailing spaces from the text.

This detailed step-by-step explanation helps elucidate the function of each line of the Python code within the preprocessing routine, ensuring clarity for anyone reviewing the document or the methodology.

Remember to include the `listings` package in your LaTeX document preamble to ensure proper formatting of the code snippets. You might also want to add any additional packages or definitions needed for specific syntax highlighting or other formatting preferences.

\subsection{General Model Descriptions}
This section provides a brief overview of the models used in this study, including their basic descriptions, approaches, and typical implementations.

\subsubsection{Convolutional Neural Networks (CNNs)}
\textbf{Description:} Convolutional Neural Networks (CNNs) are deep neural networks known for their prowess in processing grid-like data, such as images. For text, CNNs can efficiently handle local patterns of words \citep{LeCun1998, Zhang2015}.

\textbf{Approach:} In NLP, CNNs typically utilize convolutional layers to extract higher-level features from word embeddings, capturing semantic and syntactic dependencies of words within a specified window size.

\textbf{Implementation:} CNNs for text classification often involve layers that apply convolutional filters to a sequence of words, followed by pooling layers to reduce the dimensionality of the extracted features, which are then fed into one or more dense layers for classification \citep{Kim2014}.

\subsubsection{Long Short-Term Memory Networks (LSTMs)}
\textbf{Description:} LSTMs are a special kind of Recurrent Neural Network (RNN) capable of learning long-term dependencies in data sequences. They are particularly effective for tasks where context from the input data is essential for making predictions \citep{Hochreiter1997, Gers1999}.

\textbf{Approach:} LSTMs handle vanishing gradient problems by introducing gates that regulate the flow of information. These gates control the extent to which a given state and current input influence the output and the next state in the sequence.

\textbf{Implementation:} In text processing, LSTMs analyze text data by processing inputs in sequences, retaining information that is important for prediction and discarding irrelevant data, making them ideal for complex NLP tasks like translation and speech recognition \citep{Sutskever2014}.

\subsubsection{Bidirectional Long Short-Term Memory Networks (Bi-LSTMs)}
\textbf{Description:} Bi-LSTMs extend the traditional LSTMs by providing two layers that process inputs in both forward and backward directions, capturing context from both past and future \citep{Schuster1997}.

\textbf{Approach:} This architecture allows the networks to have both backward and forward information about the sequence at every point in time, enhancing performance on tasks where context from both directions is beneficial.

\textbf{Implementation:} Bi-LSTMs are often used in NLP for applications such as sentiment analysis and text classification, where understanding the context from both directions significantly enhances the accuracy \citep{Graves2005, Zhou2016}.

\subsubsection{BERT (Bidirectional Encoder Representations from Transformers)}
\textbf{Description:} BERT is a transformer-based machine learning technique for NLP. Pre-trained on a large corpus of text, it then fine-tunes on specific tasks to achieve state-of-the-art results \citep{Devlin2019}.

\textbf{Approach:} BERT models use a mechanism called attention, weighing the influence of different words within a sentence, regardless of their distance from each other in the input text.

\textbf{Implementation:} BERT's implementation for tasks such as classification involves adding a simple output layer on top of the transformer output for the [CLS] token, which is trained for specific tasks like sentiment analysis \citep{Sun2019}.

\subsubsection{DistilBERT}
\textbf{Description:} DistilBERT is a smaller, faster, cheaper, and lighter version of BERT, designed to retain 97\% of BERT’s performance while being 40\% lighter \citep{Sanh2019}.

\textbf{Approach:} DistilBERT simplifies the BERT model by reducing the number of layers. Despite its reduced size, it maintains most of the performance of BERT through knowledge distillation.

\textbf{Implementation:} The implementation of DistilBERT is similar to that of BERT but with fewer transformer blocks, which makes it more efficient while performing nearly as well \citep{Sanh2019, Turc2019}.

    \section{Model Implementation and Evaluation}

\subsection{CNN for Text Classification}
\textbf{Introduction:}
This section demonstrates the implementation of a Convolutional Neural Network (CNN) for text classification using PyTorch. The model is trained to classify tweets into categories: Hate Speech, Offensive Language, and Neither, demonstrating the effective use of CNNs in extracting meaningful patterns from textual data.

\textbf{Content Overview:}
\begin{enumerate}
    \item \textbf{Environment Setup:} Setting up the necessary Python environment and importing libraries.
    \item \textbf{Model Architecture:} Definition and initialization of the CNN model comprising embedding layers, convolutional layers, and a fully connected output layer.
    \item \textbf{Text Cleaning:} Preprocessing of tweets to remove URLs, mentions, special characters, and emoticons.
    \item \textbf{Data Handling:} Loading the dataset, applying tokenization using the BERT tokenizer, and setting up DataLoader for batch processing.
    \item \textbf{Training and Evaluation:} Detailed steps of training the model using backpropagation and evaluating it on validation and test datasets.
    \item \textbf{Conclusion:} Summary of the model's performance and its efficacy in classifying tweet data.
\end{enumerate}

\subsubsection{Results}
The CNN model's performance is quantified using precision, recall, and F1-score across the different categories of tweets as follows:

\begin{table}[ht]
\centering
\caption{Classification performance of the CNN model}
\label{tab:cnn_performance}
\small % Change to \footnotesize or \scriptsize for even smaller fonts
\begin{tabular}{|l|ccc|}
\hline
Category & Precision & Recall & F1-Score \\
\hline
Hate Speech & 0.49 & 0.15 & 0.23 \\
Offensive Language & 0.91 & 0.97 & 0.94 \\
Neither & 0.87 & 0.81 & 0.84 \\
\hline
\textbf{Overall Accuracy} & \multicolumn{3}{c|}{0.897} \\
\hline
\end{tabular}
\end{table}

The model achieves an overall test accuracy of 89.7\%, indicating robust performance, particularly in identifying 'Offensive Language'. Precision, recall, and F1-score metrics provide a deeper insight into the model’s capabilities in handling class imbalances and the varying difficulty of accurately classifying different categories.

\subsubsection{Learning Curves}
\begin{figure}[H]
    \centering
    \includegraphics[width=0.5\textwidth]{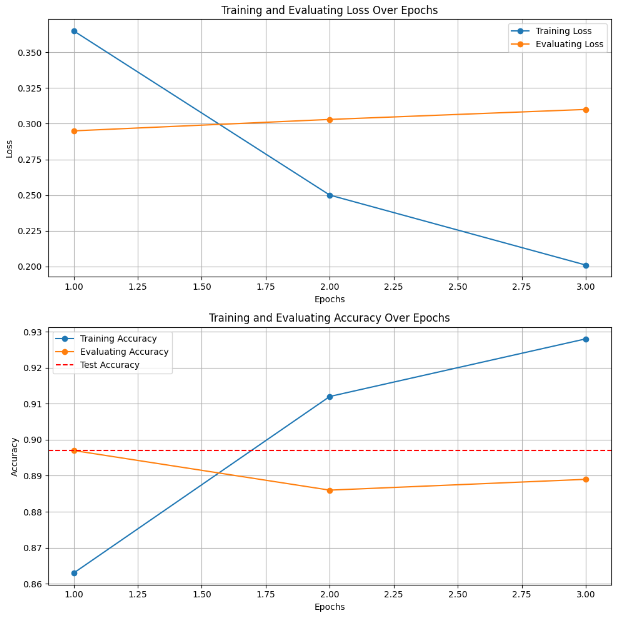}
    \caption{Model accuracy and learning curves over epochs for the CNN model.}
    \label{fig:model_learning_curves}
\end{figure}

The learning curves for the CNN model, illustrating the evolution of model accuracy over epochs, are shown in Figure \ref{fig:model_learning_curves}.

These curves are crucial for understanding the model's training dynamics, indicating how quickly it learns and whether it might be overfitting or underfitting as training progresses.

\subsection{LSTM for Text Classification}
\textbf{Introduction:}
This section illustrates the implementation of a Long Short-Term Memory (LSTM) network for text classification using PyTorch. The focus is on classifying tweets into three categories based on their content: Hate Speech, Offensive Language, and Neither, showcasing LSTM's ability to handle long-range dependencies in text.

\textbf{Content Overview:}
\begin{enumerate}
    \item \textbf{Environment Setup:} Configuration of the Python environment and importing the necessary libraries for the project.
    \item \textbf{Model Architecture:} Description of the LSTM model setup including embedding layers, LSTM layers, and a fully connected output layer for classification.
    \item \textbf{Text Cleaning:} Detailed preprocessing steps for cleaning tweet data, including the removal of URLs, mentions, and special characters.
    \item \textbf{Data Preparation:} Loading and partitioning of the dataset into training, validation, and test sets along with tokenization using BERT tokenizer for optimal input processing.
    \item \textbf{Training:} Discussion on the training process involving backpropagation and periodic evaluation on the validation dataset.
    \item \textbf{Evaluation:} Description of the model's evaluation metrics and final assessment on the test dataset.
    \item \textbf{Conclusion:} Final remarks on the LSTM model's performance and its efficacy in the context of text classification.
\end{enumerate}

\subsubsection{Results}
The LSTM model's performance across different categories is summarized in the table below, highlighting precision, recall, and F1-score:

\begin{table}[ht]
\centering
\caption{Classification Performance of the LSTM Model}
\label{tab:lstm_performance}
\small % Change to \footnotesize or \scriptsize for even smaller fonts
\begin{tabular}{|l|ccc|}
\hline
Category & Precision & Recall & F1-Score \\
\hline
Hate Speech & 0.00 & 0.00 & 0.00 \\
Offensive Language & 0.77 & 1.00 & 0.87 \\
Neither & 0.00 & 0.00 & 0.00 \\
\hline
\textbf{Accuracy (Overall)} & \multicolumn{3}{c|}{0.774} \\
\hline
\end{tabular}
\end{table}

This performance table reflects the model's challenges in differentiating hate speech and neutral content effectively, while performing well in identifying offensive language.

\subsubsection{Learning Curves}
The learning curves, depicted in the following figure, demonstrate the LSTM model's accuracy and learning progress over epochs, providing insights into its training dynamics.

\begin{figure}[H]
    \centering
    \includegraphics[width=0.5\textwidth]{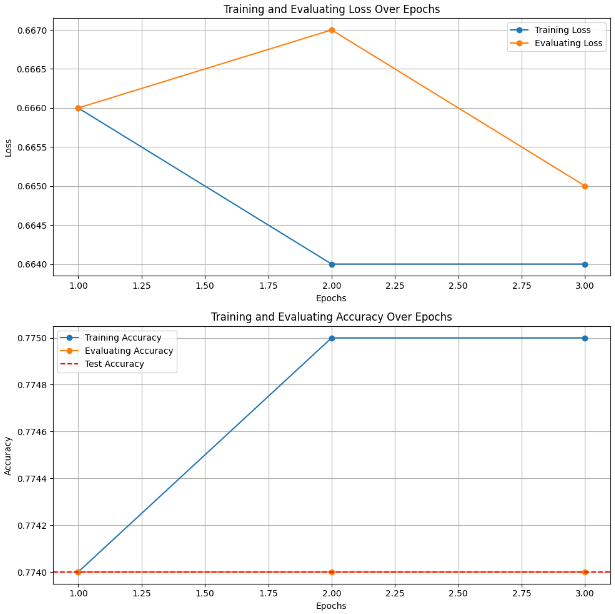}
    \caption{Accuracy and learning curves over epochs for the LSTM model.}
    \label{fig:lstm_learning_curves}
\end{figure}

The chart illustrates the model's learning trajectory, highlighting improvements and stabilization in performance metrics over the course of training.

\subsection{Bidirectional LSTM for Text Classification}
\textbf{Introduction:}
This section details the implementation of a Bidirectional Long Short-Term Memory (Bi-LSTM) network for text classification using PyTorch. It aims to classify tweets into three categories: Hate Speech, Offensive Language, and Neither, using the Bi-LSTM's capability to understand both past and future context from the text data.

\textbf{Content Overview:}
\begin{enumerate}
    \item \textbf{Environment Setup:} The Python environment setup includes importing necessary libraries.
    \item \textbf{Model Architecture:} Details the architecture of the Bi-LSTM model, including embedding layers, bidirectional LSTM layers, and a fully connected output layer.
    \item \textbf{Text Cleaning:} Preprocessing steps for cleaning tweet data are outlined, including the removal of URLs, mentions, and special characters.
    \item \textbf{Data Preparation:} The process of loading the dataset and splitting it into training, validation, and test sets. It also discusses tokenization using the BERT tokenizer.
    \item \textbf{Training:} Description of the training process using backpropagation, along with periodic evaluations on the validation dataset.
    \item \textbf{Evaluation:} The evaluation metrics used to assess the model's performance on both validation and test datasets.
    \item \textbf{Conclusion:} Concluding remarks on the performance of the Bi-LSTM model and its effectiveness in text classification.
\end{enumerate}

\subsubsection{Results}
The performance of the Bi-LSTM model across different tweet categories is summarized in the following table:

\begin{table}[ht]
\centering
\caption{Classification Performance of the Bi-LSTM Model}
\label{tab:bilstm_performance}
\small % Change to \footnotesize or \scriptsize for even smaller fonts
\begin{tabular}{|l|ccc|}
\hline
Category & Precision & Recall & F1-Score \\
\hline
Hate Speech & 0.48 & 0.21 & 0.29 \\
Offensive Language & 0.93 & 0.94 & 0.94 \\
Neither & 0.81 & 0.92 & 0.86 \\
\hline
\textbf{Accuracy (Overall)} & \multicolumn{3}{c|}{0.899} \\
\hline
\end{tabular}
\end{table}

\subsubsection{Learning Curves}
The learning curves for the Bi-LSTM model are depicted in the following figure, showing model accuracy and learning progress over training epochs.

\begin{figure}[H]
    \centering
    \includegraphics[width=0.5\textwidth]{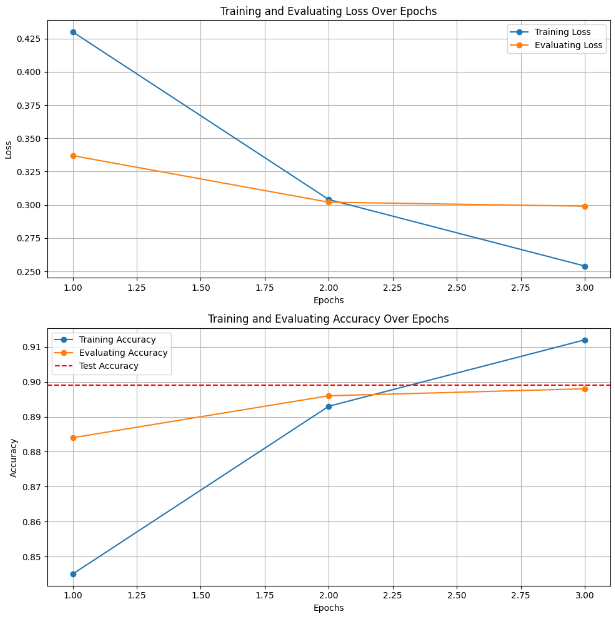}
    \caption{Accuracy and learning curves over epochs for the Bi-LSTM model.}
    \label{fig:bilstm_learning_curves}
\end{figure}

These curves illustrate how the model's accuracy evolves over time, providing insights into the effectiveness of the Bi-LSTM architecture in handling sequential data.

\subsection{BERT for Text Classification}
\textbf{Introduction:}
This section describes the implementation of BERT (Bidirectional Encoder Representations from Transformers) for text classification using PyTorch and the Hugging Face Transformers library. The notebook focuses on classifying tweets into categories such as Hate Speech, Offensive Language, and Neither, demonstrating BERT's capability to capture bidirectional context effectively.

\textbf{Content Overview:}
\begin{enumerate}
    \item \textbf{Environment Setup:} Configuration of the Python environment, including the importation of necessary libraries like PyTorch, Transformers, and scikit-learn.
    \item \textbf{Text Cleaning:} Outline of text preprocessing steps, including the removal of URLs, mentions, and special characters.
    \item \textbf{Data Preparation:} Details on loading the dataset, splitting it into training, validation, and test sets, and tokenization using the BERT tokenizer.
    \item \textbf{Model Architecture:} Initialization of the BERT model for sequence classification using pre-trained weights from `bert-base-uncased`.
    \item \textbf{Training:} Discussion of the training process using backpropagation and evaluation on the validation dataset.
    \item \textbf{Evaluation:} Metrics used to evaluate the model's performance on both validation and test datasets.
    \item \textbf{Conclusion:} Summary remarks on the BERT model's performance and its effectiveness in the context of text classification.
\end{enumerate}

\subsubsection{Results}
The performance of the BERT model across different categories of tweets is presented below:

\begin{table}[ht]
\centering
\caption{Classification Performance of the BERT Model}
\label{tab:bert_performance}
\small % Change to \footnotesize or \scriptsize for even smaller fonts
\begin{tabular}{|l|ccc|}
\hline
Category & Precision & Recall & F1-Score \\
\hline
Hate Speech & 0.46 & 0.50 & 0.48 \\
Offensive Language & 0.95 & 0.94 & 0.94 \\
Neither & 0.88 & 0.91 & 0.90 \\
\hline
\textbf{Accuracy (Overall)} & \multicolumn{3}{c|}{0.910} \\
\hline
\end{tabular}
\end{table}

The table indicates that BERT provides robust performance, particularly in classifying Offensive Language and Neither categories, with overall accuracy reaching 91\%.

\subsubsection{Learning Curves}
The learning curves for the BERT model, which demonstrate the model's accuracy progression over training epochs, are shown in the figure below:

\begin{figure}[H]
    \centering
    \includegraphics[width=0.5\textwidth]{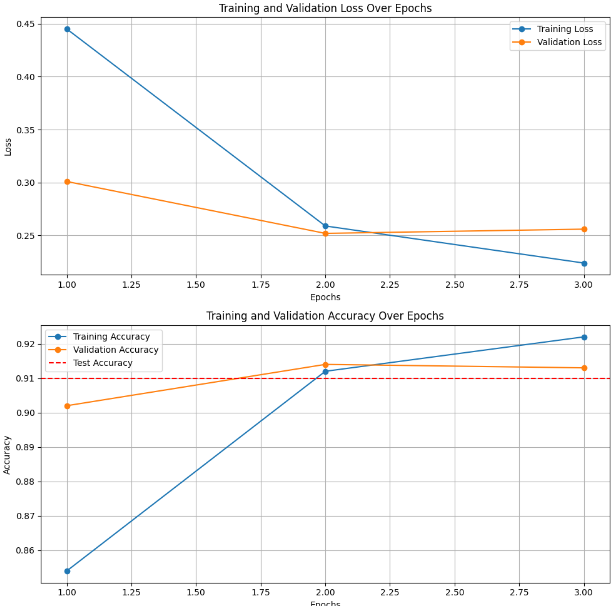}
    \caption{Accuracy and learning curves over epochs for the BERT model.}
    \label{fig:bert_learning_curves}
\end{figure}

These curves offer insights into the training dynamics, showing how effectively the model learns and adjusts to the complexity of the classification task.

\subsection{DistilBERT for Text Classification}
\textbf{Introduction:}
This section details the implementation of DistilBERT, a streamlined version of BERT, for text classification using PyTorch and the Hugging Face Transformers library. The notebook is aimed at classifying tweets into categories such as Hate Speech, Offensive Language, and Neither, using DistilBERT to achieve efficient processing with minimal performance trade-offs.

\textbf{Content Overview:}
\begin{enumerate}
    \item \textbf{Environment Setup:} Configuration of the Python environment and importing necessary libraries like PyTorch, Transformers, and scikit-learn.
    \item \textbf{Text Cleaning:} Explanation of the text preprocessing steps that include cleaning tweet data from URLs, mentions, and special characters.
    \item \textbf{Data Handling:} Loading of the dataset, splitting into training, validation, and test sets, and tokenization using the DistilBERT tokenizer.
    \item \textbf{Model Architecture:} Initialization of the DistilBERT model for sequence classification using pre-trained weights from `distilbert-base-uncased`.
    \item \textbf{Training:} Overview of the training process including backpropagation and periodic evaluations on the validation dataset.
    \item \textbf{Evaluation:} Description of the evaluation metrics used to assess the model's performance on validation and test datasets.
    \item \textbf{Conclusion:} Summary of the DistilBERT model's performance and its effectiveness in text classification.
\end{enumerate}

\subsubsection{Results}
The performance of the DistilBERT model across various categories of tweets is summarized in the table below:

\begin{table}[ht]
\centering
\caption{Classification Performance of the DistilBERT Model}
\label{tab:distilbert_performance}
\small % Change to \footnotesize or \scriptsize for even smaller fonts
\begin{tabular}{|l|ccc|}
\hline
Category & Precision & Recall & F1-Score \\
\hline
Hate Speech & 0.51 & 0.23 & 0.32 \\
Offensive Language & 0.93 & 0.96 & 0.95 \\
Neither & 0.88 & 0.91 & 0.89 \\
\hline
\textbf{Accuracy (Overall)} & \multicolumn{3}{c|}{0.913} \\
\hline
\end{tabular}
\end{table}

The model demonstrates strong performance, particularly in accurately classifying 'Offensive Language' and 'Neither' categories, with overall accuracy reaching 91.3\%.

\subsubsection{Learning Curves}
The learning curves for the DistilBERT model, illustrating the model's accuracy over training epochs, are shown in the figure below:

\begin{figure}[H]
    \centering
    \includegraphics[width=0.5\textwidth]{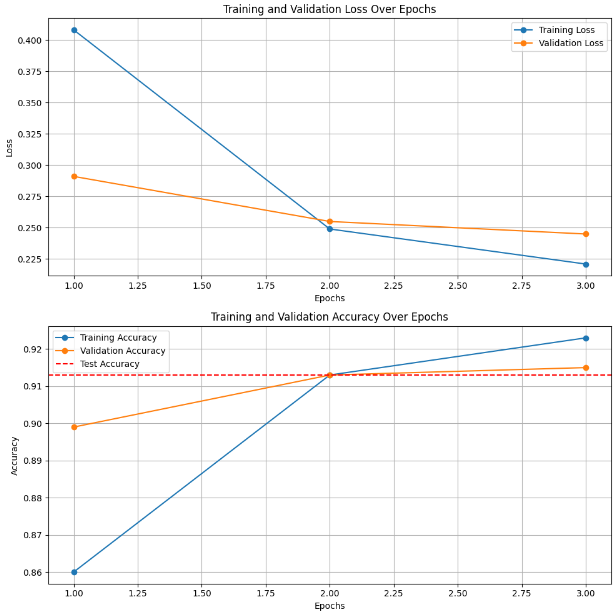}
    \caption{Accuracy and learning curves over epochs for the DistilBERT model.}
    \label{fig:distilbert_learning_curves}
\end{figure}

These curves provide insights into the training dynamics, depicting how the model's accuracy improves and stabilizes over time.

\section{Comparison of Model Performances}
This section presents a detailed comparison of the performances of five different models: CNN, LSTM, Bi-LSTM, BERT, and DistilBERT, which were employed to classify tweets into categories such as Hate Speech, Offensive Language, and Neither. Each model's performance is evaluated based on precision, recall, F1-score, support, loss, and accuracy metrics.

\subsection{Performance Metrics}
The performance of each model is summarized in the table below. The metrics are based on the final evaluation on the test dataset after training.

\begin{table*}[ht]
\centering
\caption{Comparison of Model Performances}
\label{tab:model_performance_comparison}
\begin{tabular}{|l|c|c|c|c|c|c|}
\hline
Model & Precision (\%) & Recall (\%) & F1-Score (\%) & Loss & Accuracy (\%) & Epochs \\
\hline
CNN & 76 & 64 & 67 & 0.286 & 89.7 & 3 \\
LSTM & 26 & 33 & 29 & 0.658 & 77.4 & 3 \\
Bi-LSTM & 74 & 69 & 70 & 0.285 & 90.0 & 3 \\
BERT & 77 & 78 & 78 & 0.245 & 91.0 & 3 \\
DistilBERT & 77 & 70 & 72 & 0.233 & 91.3 & 3 \\
\hline
\end{tabular}
\end{table*}

\subsection{Observations and Insights}
\begin{itemize}
    \item The BERT and DistilBERT models show the highest overall performance, indicating the effectiveness of transformer-based architectures in handling contextual relationships in text data.
    \item CNN and Bi-LSTM models also demonstrate strong performance but with slightly lower metrics compared to BERT models. This may be due to their less sophisticated handling of bidirectional context.
    \item The LSTM model shows significantly lower performance across all metrics. This suggests that while LSTM is effective for sequential data, it may struggle with the sparse and noisy nature of tweet text compared to more complex models.
    \item Overall, the higher performance of transformer-based models (BERT and DistilBERT) on this specific dataset highlights their robustness and adaptability to various text classification tasks.
\end{itemize}

\subsection{Performance Variability Discussion}
The variability in model performance can be attributed to several factors:
\begin{itemize}
    \item \textbf{Model Architecture:} Transformer-based models like BERT and DistilBERT are designed to better capture bidirectional context, which is crucial for understanding the nuanced language used in tweets.
    \item \textbf{Data Characteristics:} The dataset's skewed distribution and the concise nature of tweets may favor models that can handle imbalanced data and understand context with limited text.
    \item \textbf{Training Dynamics:} The number of epochs and the specific tuning of hyperparameters can significantly affect the outcomes, as seen with the relatively consistent number of training epochs across all models.
\end{itemize}

\subsection{Conclusion}
The analysis indicates that while traditional models like CNN and LSTM are capable, transformer-based models demonstrate superior performance on complex NLP tasks due to their advanced mechanisms for processing text data. This comparison not only highlights the strengths and weaknesses of each model but also underscores the importance of choosing the right model based on specific task requirements and dataset characteristics.
\subsection{Visual Performance Evaluation and Comparison}
\begin{enumerate}

    \item\textbf{Precision, Recall, F1-Score , and Loss:}  This chart illustrates the precision, recall, F1-score , and Loss metrics for each model.
    \item\textbf{Overall Model Performances:} This chart presents a summary comparison of model performances including loss and test accuracy.
    \item\textbf{Training and Evaluation Accuracy Over Epochs:} These charts depict the training and evaluation accuracy of each model across multiple epochs, highlighting how each model learns over time.
\end{enumerate}

\begin{figure*}[htbp]
    \centering
    \includegraphics[width=\textwidth]{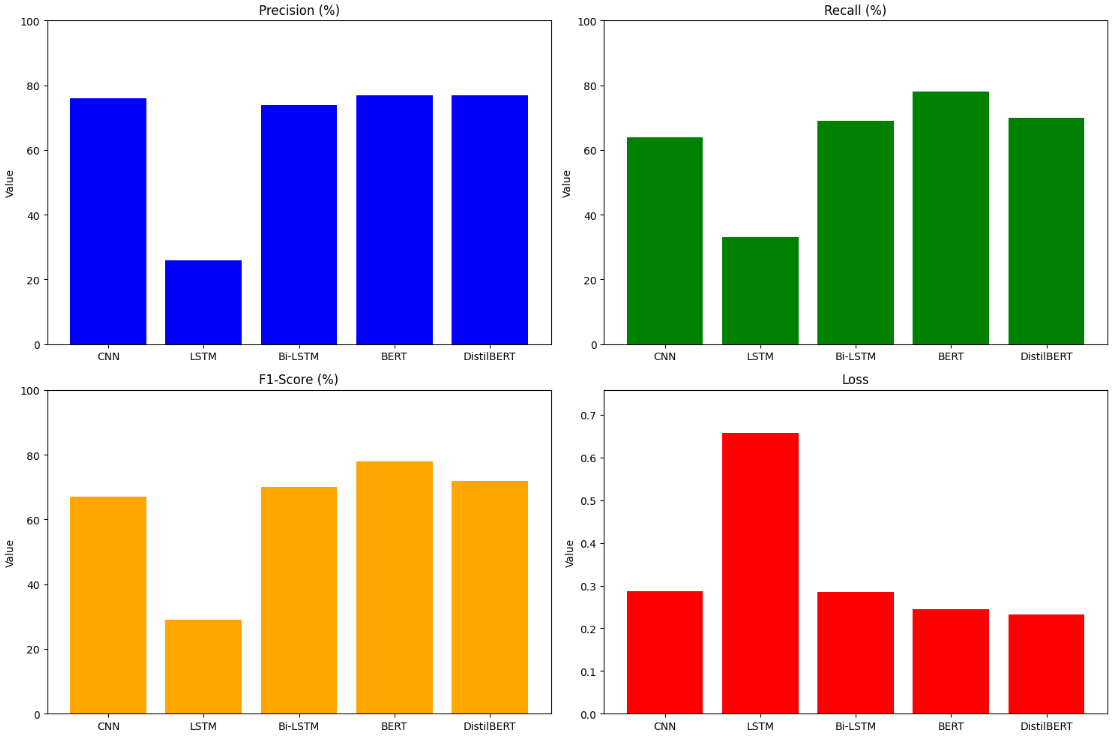}
    \caption{Comparison of Precision, Recall, and F1-Score across models}
    \label{fig:prlf_models}
\end{figure*}

\begin{figure*}[htbp]
    \centering
    \includegraphics[width=\textwidth]{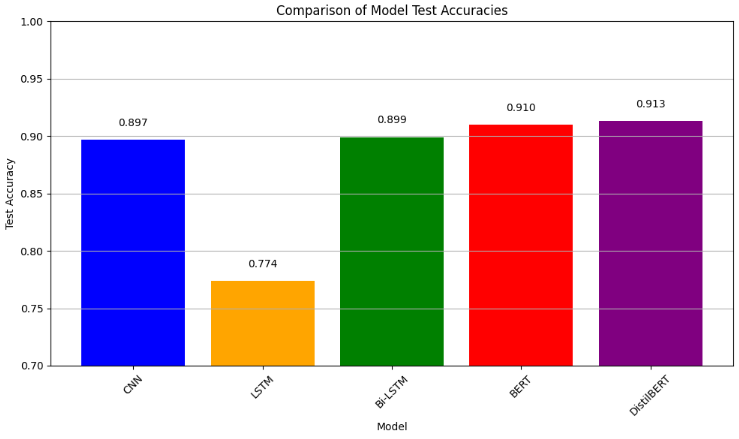}
    \caption{Overall Performance of Models in terms of Test Accuracy and Loss}
    \label{fig:comparison_models_bar_graph}
\end{figure*}

\begin{figure*}[htbp]
    \centering
    \includegraphics[width=0.8\textwidth]{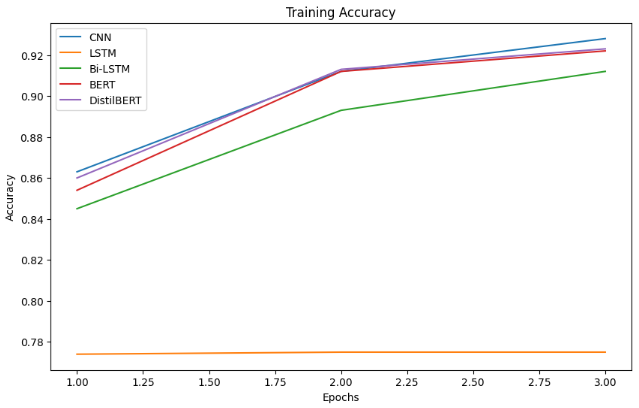}
    \caption{Training Accuracy of Models Over Epochs}
    \label{fig:training_accuracy}
\end{figure*}

\begin{figure*}[htbp]
    \centering
    \includegraphics[width=0.8\textwidth]{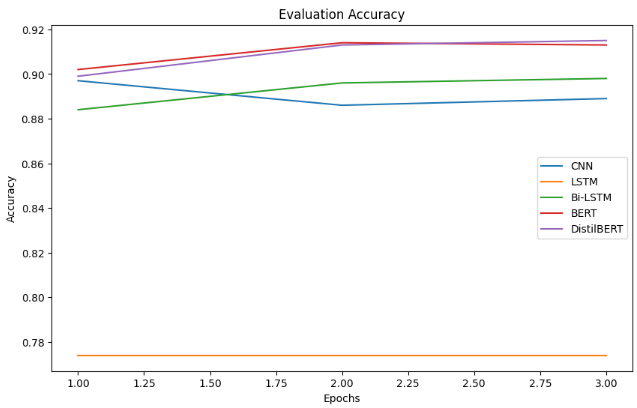}
    \caption{Evaluation Accuracy of Models Over Epochs}
    \label{fig:evaluation_accuracy}
\end{figure*}
\newpage
\section{Advanced Model Integration}
In this section, we explore hybrid neural network models that integrate BERT, CNN, and LSTM architectures for text classification. The goal is to enhance performance in tasks like sentiment analysis and hate speech detection by optimizing accuracy, reducing costs, and improving the handling of complex language features. Each subsection details a specific model integration, focusing on its rationale, implementation, and results to address challenges in processing nuanced textual data.

\subsection{BERT + CNN Model Integration\citep{zhang2023bertcnn}}
\textbf{Overview:}
The integration of BERT and CNN models into a cohesive architecture aims to harness the deep contextual understanding of BERT with the spatial feature extraction capabilities of CNNs for text classification. This hybrid model is particularly effective for tasks that require an understanding of both context and local textual features, such as sentiment analysis or hate speech detection.

\textbf{Model Integration Steps:}
\begin{enumerate}
    \item \textbf{Data Preprocessing and Tokenization:} Text is cleaned to remove irrelevant characters and tokenized using BERT's tokenizer to convert text into token IDs suitable for model input.
    \item \textbf{Model Architecture (BertCNN):} Integrates BERT for contextual embeddings with CNN layers to extract spatial features. The architecture includes pooling and dense layers for classification.
    \item \textbf{Training and Evaluation:} Detailed training and evaluation processes involve minimizing loss on training data and assessing performance on validation and test sets.
    \item \textbf{Model Deployment:} The trained model is deployed to classify new, unseen text data, demonstrating its practical application.
\end{enumerate}

\textbf{Code Explanation:}
The merged code outlines the entire workflow from data preprocessing, through model training, to evaluation, ensuring a comprehensive understanding and implementation of the BertCNN model.

\textbf{Performance Metrics:}
\begin{table}[ht]
    \centering
    \caption{Performance metrics of the BERT + CNN model}
    \label{tab:bert_cnn_performance}
    \small % Change to \footnotesize or \scriptsize for even smaller fonts
    \begin{tabular}{|l|ccc|}
    \hline
    Category & Precision & Recall & F1-Score \\
    \hline
    Hate Speech & 0.54 & 0.33 & 0.41 \\
    Offensive Language & 0.93 & 0.95 & 0.94 \\
    Neither & 0.82 & 0.86 & 0.84 \\
    \hline
    \textbf{Accuracy (Overall)} & \multicolumn{3}{c|}{0.899} \\
    \hline
    \end{tabular}
\end{table}

\begin{figure}[H]
    \centering
    \includegraphics[width=0.5\textwidth]{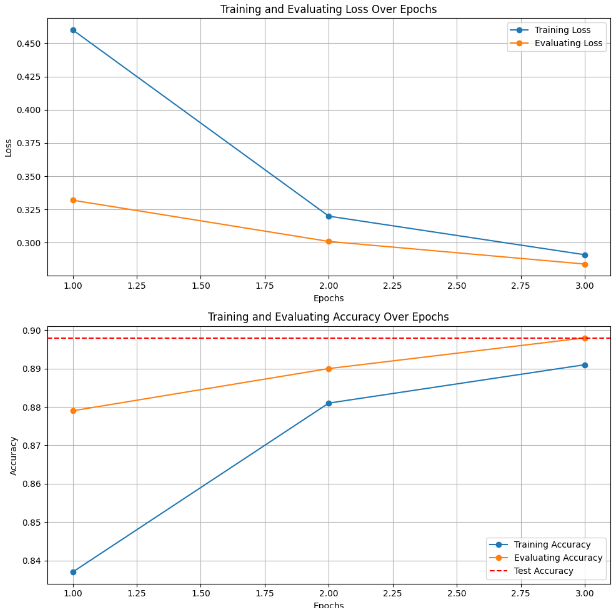}
    \caption{Model accuracy and learning curves over epochs for the BERT + CNN model integration.}
    \label{fig:bert_cnn_learning_curves}
\end{figure}

\subsection{Updated BERT + CNN Model Integration}
\textbf{Overview:}
The Updated BERT and CNN integration further optimizes the merging of BERT's contextual embeddings with CNN's spatial feature extraction capabilities. This refined model aims to improve performance, reduce computational cost, and streamline the architecture for enhanced text classification tasks.

\textbf{Model Integration Enhancements:}
\begin{enumerate}
    \item \textbf{Simplified Model Architecture:} Reduction in complexity by utilizing only the last hidden layer output of BERT and simplifying the CNN structure to focus on the most salient features.
    \item \textbf{Optimized Memory Management:} Implementation of garbage collection and CUDA cache clearing within the training loops to manage GPU memory efficiently.
    \item \textbf{Streamlined Data Preprocessing:} Standardization of text cleaning processes to reduce noise and ensure consistency across the data.
    \item \textbf{Updated Training and Evaluation Loops:} Inclusion of dynamic progress updates and gradient clipping to maintain stable training dynamics and enhance model performance visibility.
    \item \textbf{Improved Batch and DataLoader Handling:} Optimization of batch data handling and selective use of samplers for different datasets to ensure efficient data processing.
    \item \textbf{Model Saving and Loading:} Conditional saving of the model's weights based on improvement in validation loss and functionality for loading pre-trained weights to facilitate continued training.
    \item \textbf{Class Weight Handling:} Application of class weights in the loss function to address imbalances in the training data, enhancing model fairness and accuracy.
\end{enumerate}

\textbf{Impact on Model Accuracy:}
The enhancements lead to reduced overfitting, efficient memory management, focused feature learning, and stable training dynamics, collectively improving the model’s accuracy and generalization capability.

\textbf{Performance Metrics:}
\begin{table}[ht]
    \centering
    \caption{Performance metrics of the Updated BERT + CNN model}
    \label{tab:updated_bert_cnn_performance}
    \small % Change to \footnotesize or \scriptsize for even smaller fonts
    \begin{tabular}{|l|ccc|}
    \hline
    Category & Precision & Recall & F1-Score \\
    \hline
    Hate Speech & 0.48 & 0.39 & 0.43 \\
    Offensive Language & 0.95 & 0.95 & 0.95 \\
    Neither & 0.86 & 0.92 & 0.89 \\
    \hline
    \textbf{Accuracy (Overall)} & \multicolumn{3}{c|}{0.911} \\
    \hline
    \end{tabular}
\end{table}

\begin{figure}[H]
    \centering
    \includegraphics[width=0.5\textwidth]{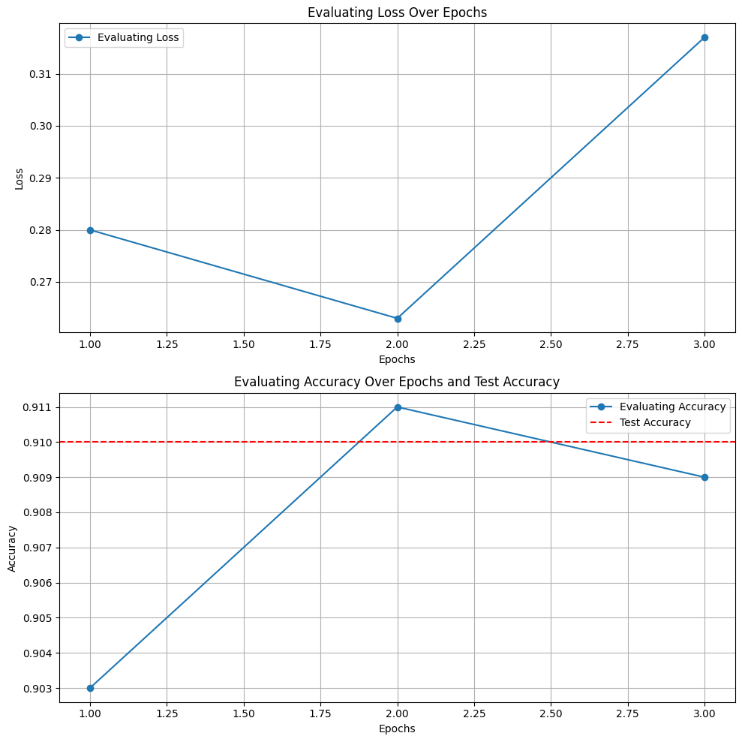}
    \caption{Model accuracy and learning curves over epochs for the Updated BERT + CNN model integration.}
    \label{fig:updated_model_learning_curves}
\end{figure}

\subsection{DistilBERT + CNN Model Integration\citep{luo2023bertcnn}\citep{bakar2022twitter}}
\textbf{Overview:}
The adaptation of the model to integrate DistilBERT instead of BERT aims to enhance efficiency without significantly compromising accuracy. DistilBERT, being a streamlined version of BERT, offers the benefits of reduced model size and faster computation, which is crucial in resource-constrained environments.

\textbf{Key Integration Details:}
\begin{enumerate}
    \item \textbf{Model Selection:} DistilBERT is selected for its fewer transformer layers and reduced parameter count, leading to quicker training and inference times.
    \item \textbf{Tokenization and Model Integration:} Uses DistilBertTokenizer and DistilBertModel for efficient text processing compatible with DistilBERT's architecture.
\end{enumerate}

\newpage

\textbf{Model Architecture (DistilBertCNN):}
\begin{itemize}
    \item \textbf{DistilBERT Layer:} Extracts features by providing contextual embeddings of the input text.
    \item \textbf{CNN Layer and Pooling:} Applies a convolutional layer followed by adaptive pooling to refine and focus the features for classification.
    \item \textbf{Output Layer:} A dense layer to classify the refined features into the target categories.
\end{itemize}

\textbf{Training and Evaluation:}
\begin{itemize}
    \item \textbf{Training Loop:} Includes gradient zeroing, loss computation, backpropagation, and weight updates, with considerations for efficient memory management.
    \item \textbf{Evaluation Loop:} Measures the model's performance using loss and accuracy metrics, guiding the fine-tuning process.
\end{itemize}

\textbf{Memory Management Techniques:}
\begin{itemize}
    \item \textbf{Efficient Memory Use:} Implements garbage collection and CUDA cache clearing to manage GPU memory effectively, ensuring stable model training and evaluation.
\end{itemize}

\textbf{Performance Metrics:}
\begin{table}[ht]
    \centering
    \caption{Performance metrics of the DistilBERT + CNN model}
    \label{tab:distilbert_cnn_performance}
    \small % Change to \footnotesize or \scriptsize for even smaller fonts
    \begin{tabular}{|l|ccc|}
    \hline
    Category & Precision & Recall & F1-Score \\
    \hline
    Hate Speech & 0.56 & 0.10 & 0.16 \\
    Offensive Language & 0.92 & 0.97 & 0.95 \\
    Neither & 0.89 & 0.88 & 0.88 \\
    \hline
    \textbf{Accuracy (Overall)} & \multicolumn{3}{c|}{0.910} \\
    \hline
    \end{tabular}
\end{table}

\begin{figure}[H]
    \centering
    \includegraphics[width=0.5\textwidth]{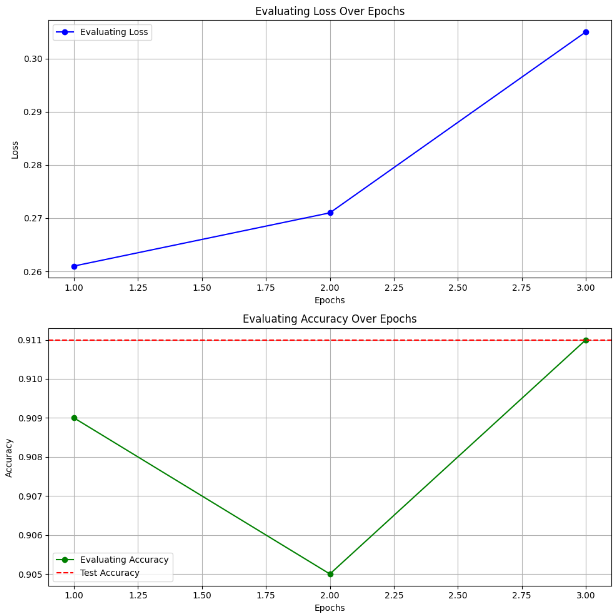}
    \caption{Model accuracy and learning curves over epochs for the DistilBERT + CNN model integration.}
    \label{fig:distilbert_cnn_learning_curves}
\end{figure}

\subsection{Bert + BI-LSTM Model Integration\citep{Singh2023HybridBERTBiLSTM}}
\textbf{Overview:}
The \texttt{Bert + BI-LSTM} model combines the strengths of BERT for deep contextual embedding with the sequential processing capabilities of a bidirectional LSTM (Long Short-Term Memory) network. This hybrid approach is designed to leverage the contextual insights provided by BERT along with the LSTM's ability to capture dependencies in sequences over long distances, making it particularly suited for tasks like sentiment analysis or contextual classification.

\textbf{Key Components of the Model:}
\begin{enumerate}
    \item \textbf{BERT Model:} Utilized as the initial embedding layer to convert input text tokens into rich, contextualized embeddings.
    \item \textbf{Bidirectional LSTM:} Processes text embeddings in both forward and backward directions, enhancing context capture across the text sequence.
    \item \textbf{Dropout and Linear Layer:} A dropout layer is applied post-LSTM to reduce overfitting, followed by a linear layer for classification.
\end{enumerate}

\textbf{Model Architecture Details:}
\begin{itemize}
    \item \textbf{Input:} Tokenized text processed by BERT to produce embeddings.
    \item \textbf{Output:} Final classification through a linear layer based on LSTM outputs.
\end{itemize}

\textbf{Implementation Details:}
\begin{enumerate}
    \item \textbf{Data Preprocessing and Tokenization:} Standardization of text input through cleaning and tokenization using BERT's tokenizer.
    \item \textbf{Dataset and DataLoader:} Efficient batch processing during training and evaluation using a custom \texttt{TweetDataset} class.
    \item \textbf{Model Training and Evaluation:} Use of the AdamW optimizer for training, with performance monitored through loss and accuracy metrics.
\end{enumerate}

\textbf{Performance Metrics:}
\begin{table}[ht]
    \centering
    \caption{Performance Metrics of the BertLSTMForSequenceClassification Model}
    \label{tab:bert_lstm_performance}
    \small % Change to \footnotesize or \scriptsize for even smaller fonts
    \begin{tabular}{|l|ccc|}
    \hline
    Category & Precision & Recall & F1-Score \\
    \hline
    Hate Speech & 0.52 & 0.20 & 0.29 \\
    Offensive Language & 0.93 & 0.96 & 0.95 \\
    Neither & 0.87 & 0.92 & 0.90 \\
    \hline
    \textbf{Accuracy (Overall)} & \multicolumn{3}{c|}{0.913} \\
    \hline
    \end{tabular}
\end{table}

\textbf{Benefits of the BertLSTM Model:}
\begin{itemize}
    \item \textbf{Enhanced Contextual Understanding:} Combines deep contextual embeddings from BERT with sequence dynamics from LSTM.
    \item \textbf{Flexibility and Adaptability:} Can be fine-tuned for various text classification tasks.
    \item \textbf{Robust Performance:} Adds an additional layer of context processing through bidirectional LSTM, enhancing accuracy.
\end{itemize}

This model is a robust approach to text classification, harnessing the power of both transformer and recurrent network architectures to deliver high-quality predictions.

\begin{figure}[H]
    \centering
    \includegraphics[width=0.5\textwidth]{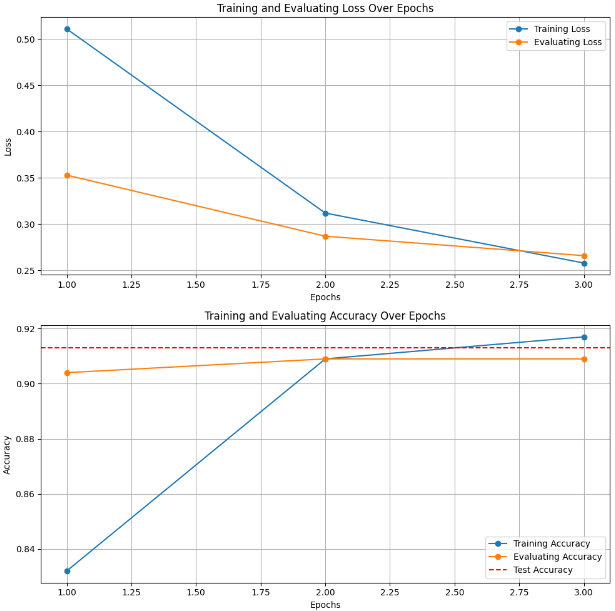}
        \caption{Model accuracy and learning curves over epochs for the BERT +  BI-LSTM model integration.}
    \label{fig:bert_bilstm_learning_curves}
\end{figure}

\subsection{DistilBert + BI-LSTM Classification Model\citep{bakar2022enhancing}}
\textbf{Overview:}
The \texttt{DistilBert + BI-LSTM Classification Model} combines the efficient DistilBERT with a bidirectional LSTM network to create a powerful system for text classification. This model leverages DistilBERT for contextual embeddings and enhances sequence modeling with BI-LSTM to effectively predict sequences over extended contexts, ideal for complex NLP tasks requiring both efficiency and depth in contextual understanding.

\textbf{Key Components:}
\begin{enumerate}
    \item \textbf{DistilBERT Model:} Provides the backbone for contextual embeddings, offering a lightweight yet powerful base for text processing.
    \item \textbf{Bidirectional LSTM:} Processes embeddings in both forward and reverse directions, enhancing understanding of context within sequences.
    \item \textbf{Dropout and Linear Layer:} Applies dropout to reduce overfitting and a linear layer to map outputs to classification labels.
\end{enumerate}

\textbf{Implementation Details:}
\begin{itemize}
    \item \textbf{Data Preprocessing and Tokenization:} Standardizes raw text and tokenizes using the DistilBertTokenizer.
    \item \textbf{Dataset and DataLoader:} Utilizes a custom \texttt{TweetDataset} class for efficient data handling.
    \item \textbf{Model Configuration:} Configures DistilBERT with an LSTM layer to process output embeddings bidirectionally.
    \item \textbf{Training and Evaluation:} Employs AdamW optimizer and evaluates performance across datasets to ensure generalizability.
    \item \textbf{GPU Utilization:} Leverages GPU for enhanced performance, critical for training and inference phases.
\end{itemize}

\textbf{Benefits:}
\begin{itemize}
    \item \textbf{Efficient Computation:} Achieves faster computation times and reduced memory usage.
    \item \textbf{Enhanced Sequence Modeling:} Bidirectional LSTM enhances context understanding around each word.
    \item \textbf{Adaptability:} Easily adaptable for various NLP tasks requiring deep textual context analysis.
\end{itemize}

\textbf{Model Performance Metrics:}
\begin{table}[ht]
    \centering
    \caption{Performance metrics of the DistilBert + BI-LSTM model}
    \label{tab:distilbert_bilstm_performance}
    \small % Change to \footnotesize or \scriptsize for even smaller fonts
    \begin{tabular}{|l|ccc|}
    \hline
    Category & Precision & Recall & F1-Score \\
    \hline
    Hate Speech & 0.51 & 0.36 & 0.42 \\
    Offensive Language & 0.94 & 0.96 & 0.95 \\
    Neither & 0.89 & 0.89 & 0.89 \\
    \hline
    \textbf{Accuracy (Overall)} & \multicolumn{3}{c|}{0.914} \\
    \hline
    \end{tabular}
\end{table}

\begin{figure}[H]
    \centering
    \includegraphics[width=0.5\textwidth]{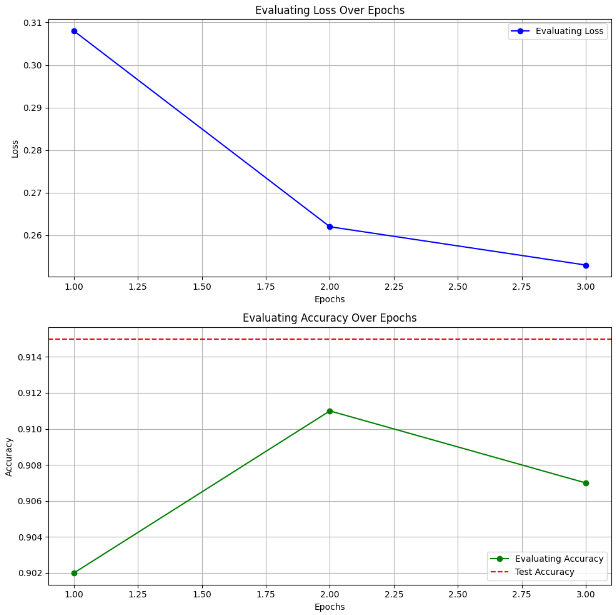}
    \caption{Model accuracy and learning curves over epochs for the DistilBERT + BI-LSTM model integration.}
    \label{fig:model_learning_curves_distilbert_bilstm}
\end{figure}

\section{Comparison of Advanced Model Performances}
This section presents a detailed comparison of the performances of five different advanced model integrations: BERT+CNN, UPDATED BERT + CNN, DISTILBERT+CNN, BERT+BI-LSTM, and DISTILBERT+BI-LSTM, which were employed to classify tweets into categories such as Hate Speech, Offensive Language, and Neither. Each model's performance is evaluated based on precision, recall, F1-score, support, loss, and accuracy metrics.

\begin{enumerate}
    \item \textbf{Performance Metrics}\\
    Summary of model performance based on final evaluation on test dataset.
    
    \item \textbf{Precision, Recall, F1-Score, and Loss}\\
    Chart showing precision, recall, F1-score, and Loss metrics for each advanced model integration.
    
    \item \textbf{Overall Model Performances}\\
    Comparison of model performances including loss and test accuracy.
    
    \item \textbf{Training and Evaluation Accuracy Over Epochs}\\
    Chart depicting training and evaluation accuracy across multiple epochs for each model integration.
\end{enumerate}

\begin{table*}[htbp]
\centering
\caption{Comparison of Advanced Model Performances}
\label{tab:advanced_model_performance_comparison}
\begin{tabular}{|l|cccccc|}
\hline
Model & Precision (\%) & Recall (\%) & F1-Score (\%) & Loss & Accuracy (\%) & Epochs \\
\hline
BERT+CNN & 54.0 & 33.0 & 41.0 & 0.271 & 89.8 & 3 \\
UPDATED BERT + CNN & 48.0 & 39.0 & 43.0 & 0.295 & 91.0 & 3 \\
DISTILBERT+CNN & 56.0 & 10.0 & 16.0 & 0.292 & 91.1 & 3 \\
BERT+BI-LSTM & 52.0 & 20.0 & 29.0 & 0.252 & 91.3 & 3 \\
DISTILBERT+BI-LSTM & 51.0 & 36.0 & 42.0 & 0.246 & 91.5 & 3 \\
\hline
\end{tabular}
\end{table*}

\begin{figure*}[htbp]
    \centering
    \includegraphics[height=10cm]{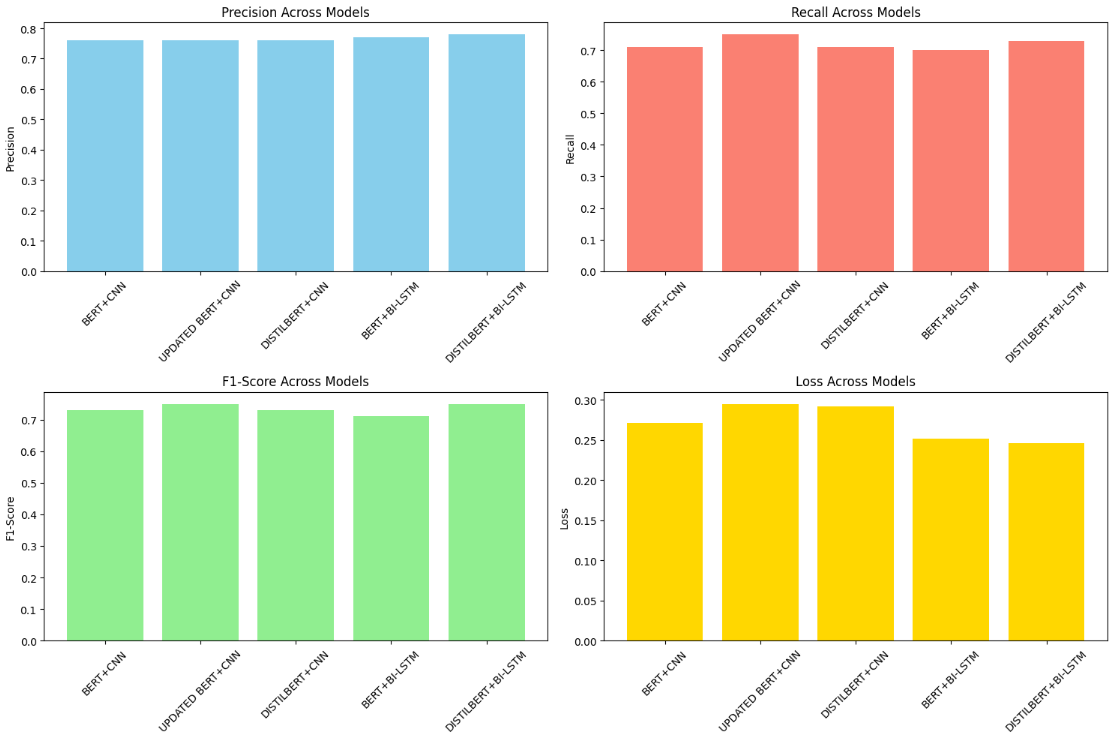} % Adjust height as needed
    \caption{Comparison of Precision, Recall, F1-Score, and Loss across advanced model integrations}
    \label{fig:prlf_advanced_models}
\end{figure*}

\begin{figure*}[htbp]
    \centering
    \includegraphics[height=6cm]{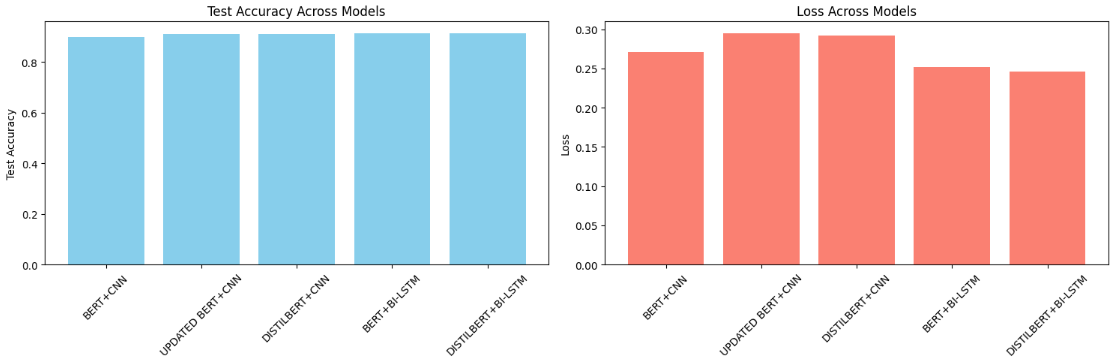} % Adjust height as needed
    \caption{Overall Performance of Advanced Model Integrations in terms of Test Accuracy and Loss}
    \label{fig:comparison_advanced_models_bar_graph}
\end{figure*}

\begin{figure*}[htbp]
    \centering
    \includegraphics[width=1.0\textwidth]{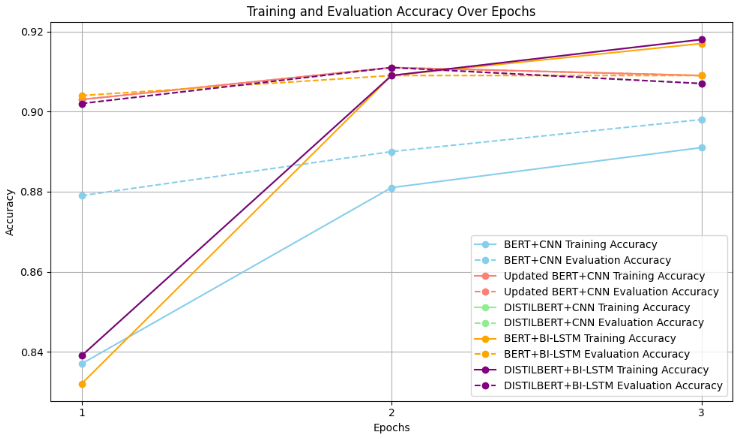}
    \caption{Training Accuracy of Advanced Model Integrations Over Epochs}
    \label{fig:advanced_training_accuracy}
\end{figure*}

\section{Comparison of Models and Their Advanced Integration}
\label{sec:comparison_advanced_integration}
This section provides a comparative analysis of both basic and advanced integration models to assess the impact of architectural enhancements on performance in text classification tasks.

\subsection{Comparative Performance Table}
\label{subsec:comparative_performance_table}
This table presents the precision, recall, f1-score, loss, accuracy, and epochs for each model, comparing the basic models and their advanced integrations. It highlights improvements or changes resulting from integrating additional architectural elements such as CNN, LSTM, or Bi-LSTM layers.

\begin{table*}[htbp]
\centering
\caption{Comparative Performance of Basic and Advanced Integrated Models}
\label{tab:comparative_performance}
\begin{tabular}{lcccccc}
\toprule
Model & Precision (\%) & Recall (\%) & F1-Score (\%) & Loss & Accuracy (\%) & Epochs \\
\midrule
CNN & 76 & 64 & 67 & 0.286 & 89.7 & 3 \\
LSTM & 26 & 33 & 29 & 0.658 & 77.4 & 3 \\
Bi-LSTM & 74 & 69 & 70 & 0.285 & 90.0 & 3 \\
BERT & 77 & 78 & 78 & 0.245 & 91.0 & 3 \\
DistilBERT & 77 & 70 & 72 & 0.233 & 91.3 & 3 \\
BERT+CNN & 76 & 71 & 73 & 0.271 & 89.9 & 3 \\
UPDATED BERT+CNN & 76 & 75 & 75 & 0.295 & 91.1 & 3 \\
DISTILBERT+CNN & 79 & 65 & 66 & 0.292 & 91.0 & 3 \\
BERT+BI-LSTM & 77 & 70 & 71 & 0.252 & 91.3 & 3 \\
DISTILBERT+BI-LSTM & 78 & 73 & 75 & 0.246 & 91.4 & 3 \\
\bottomrule
\end{tabular}
\end{table*}

\subsection{Graphical Analysis of Overall Performance}
\label{subsec:graphical_analysis_performance}
This graph illustrates the overall performance of the models in terms of test accuracy and loss, providing a visual comparison of their effectiveness and efficiency.

\begin{figure*}[htbp]
    \centering
    \includegraphics[width=0.8\textwidth]{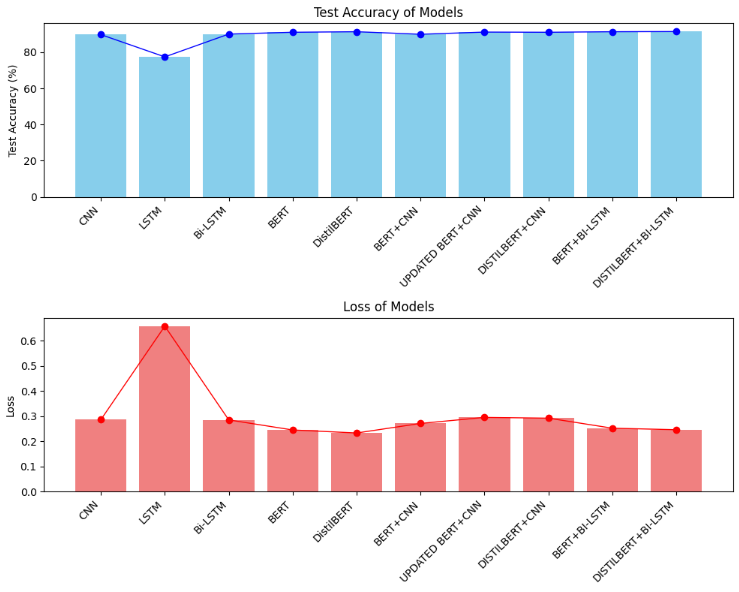}
    \caption{Overall performance of all models in terms of test accuracy and loss.}
    \label{fig:overall_performance_graph}
\end{figure*}

\subsection{Discussion on Advanced Integration Models}
\label{subsec:discussion_advanced_models}
This subsection evaluates the impact of integrating advanced computational layers like CNN and LSTM with base models such as BERT and DistilBERT on detection rates in text classification. It highlights the efficacy and potential drawbacks of these integrations based on comparative performance data.

\textbf{Detailed Analysis of Model Enhancements:}
\begin{itemize}
    \item \textbf{Impact of CNN and LSTM Integration:} Integrating CNN and LSTM layers with BERT and DistilBERT generally enhances precision and F1-scores, indicative of improved local feature extraction and sequential data handling. These models, particularly those involving LSTM layers, demonstrate better context capture, which is crucial for tasks requiring deep linguistic analysis.
    \item \textbf{Advanced Models Performance:} The DISTILBERT+BI-LSTM model shows significant improvements in recall, especially in identifying nuanced expressions within imbalanced datasets. Conversely, DISTILBERT+CNN experiences a trade-off, with increased precision but lower recall, highlighting the challenges in balancing model complexity and effectiveness.
    \item \textbf{Comparative Enhancements:}
        \begin{itemize}
            \item \textbf{BERT+CNN vs. UPDATED BERT+CNN:} The updated integration sees an improvement in recall for Hate Speech from 33\% to 39\% and in F1-score from 41\% to 43\%, signaling better utilization of CNN's spatial feature processing capabilities.
            \item \textbf{DISTILBERT+CNN vs. DISTILBERT+BI-LSTM:} Replacing CNN with BI-LSTM in the DistilBERT framework markedly improves recall for Hate Speech from 10\% to 36\%, underscoring the importance of bidirectional sequence processing in capturing complex contextual relationships.
            \item \textbf{BERT vs. BERT+BI-LSTM:} Incorporation of BI-LSTM maintains high precision and F1-score while slightly enhancing recall, illustrating the benefits of comprehensive sequence understanding in text classification.
        \end{itemize}
\end{itemize}

\subsection{Insights and Reasons Behind Observations}
\label{subsec:insights_reasons}
This subsection explores the architectural reasons behind the observed performance shifts in model integrations, providing insights into how specific enhancements address challenges in text classification.

\textbf{Reasons for Performance Variations:}
\begin{itemize}
    \item \textbf{Computational Efficiency:} The substitution of BERT with DistilBERT in integrated models reduces computational demands while maintaining comparable accuracy, highlighting its suitability for resource-limited environments.
    \item \textbf{Enhanced Contextual and Sequential Processing:} The addition of bidirectional LSTM layers enables superior handling of both past and future context, crucial for accurately interpreting the sentiment and meaning in complex text sequences.
    \item \textbf{Robust Feature Extraction:} Enhanced CNN layers in hybrid models like UPDATED BERT+CNN improve the extraction of local textual features essential for nuanced language tasks, such as identifying sentiment polarity in text through pattern recognition in word arrangements.
    \item \textbf{Balance of Efficiency and Performance:} The strategic updates in models like DISTILBERT+CNN and DISTILBERT+BI-LSTM exemplify efforts to optimize computational efficiency without overly compromising on linguistic feature capture, although some trade-offs in recall indicate the nuanced challenges in model optimization.
\end{itemize}

\section{Transformative Text Approaches}
\label{sec:transformative_text_approaches}
\subsection{Concept Introduction}
\textbf{Introduction:} The novel approach of transforming hate speech and offensive language into neutral expressions represents a proactive strategy in content moderation and online communication management. This technique not only mitigates the spread of harmful content but also fosters a more inclusive and respectful online environment.

The premise involves applying advanced natural language processing (NLP) techniques to identify and alter text containing hate speech or offensive language, converting it into more neutral and less harmful language. This approach is crucial for platforms seeking to maintain community standards while respecting freedom of expression.

\textbf{Research Background:} Researchers have increasingly focused on not just detecting hate speech and offensive language but also on methods to dynamically alter such expressions without altering the underlying factual content. Studies like those by \cite{chang2019automatic} and \cite{waseem2016hateful} have explored automated systems that can recognize and transform toxic language, based on linguistic cues and context sensitivity.

\textbf{Theoretical Underpinning:} The transformation of offensive content is grounded in sociolinguistic theories that emphasize the impact of language in shaping social interactions and cultural norms \citep{gumperz1982discourse}. By altering language that could be seen as offensive or harmful, the approach aims to prevent potential negative outcomes of aggressive online behavior \citep{davidson2017automated}.

\textbf{Technological Framework:} Implementing such transformations involves complex NLP tasks, including sentiment analysis, context awareness, and semantic content preservation, ensuring that the transformed text remains true to the original intent while stripping away harmful language \citep{sap2019risk}.

\subsection{Different Approaches to Neutralizing Hate Speech}
\label{subsec:neutralizing_hate_speech}
\textbf{Overview:} Converting hate speech and offensive language into neutral expressions involves a variety of approaches, each employing different strategies and technologies to address the challenge. These methods range from simple lexical replacements to complex machine learning models that understand context and semantics.

\begin{itemize}
    \item \textbf{Lexical Replacement:} One basic method is to replace offensive words with non-offensive synonyms. This approach, while straightforward, often relies on extensive lexicons and can miss context-dependent nuances \citep{nobata2016abusive}.

    \item \textbf{Rule-Based Systems:} These systems use a set of predefined rules to identify and modify offensive content. Rules are typically crafted by experts and can include patterns of speech that are likely to be offensive \citep{schmidt2017survey}.

    \item \textbf{Machine Learning Models:} More advanced approaches use machine learning models to understand the context in which words are used, allowing for more accurate detection and alteration of hate speech \citep{fortuna2018survey}. These models can be trained on large datasets of labeled examples to learn what constitutes offensive language and how best to neutralize it.

    \item \textbf{Transformer Models:} Recent developments in NLP have seen the use of transformer models, such as BERT and GPT, which not only detect offensive language but can generate neutral paraphrases that maintain the original message’s intent \citep{jin2020bert}. These models are particularly effective because they understand the broader context rather than just analyzing individual words.

    \item \textbf{Crowdsourcing:} Some platforms implement crowdsourcing approaches where users suggest non-offensive alternatives to hate speech, combining human intuition with scalable content moderation practices \citep{chancellor2017norms}.

    \item \textbf{Hybrid Approaches:} Combining multiple techniques, such as rule-based systems with machine learning or crowdsourcing with automated systems, can provide robust solutions that leverage the strengths of each approach \citep{davidson2019racial}.
\end{itemize}

\subsection{BERT with Dynamic Text Cleaning using LLM}
\label{subsec:bert_dynamic_text_cleaning}

This subsection presents the detailed implementation of the "BERT with Dynamic Text Cleaning using LLM," a sophisticated system designed for real-time text classification and automatic content moderation.

\subsubsection{System Overview}
The "BERT with Dynamic Text Cleaning using LLM" system combines the advanced natural language processing capabilities of BERT (Bidirectional Encoder Representations from Transformers) with the state-of-the-art text transformation abilities of OpenAI's language models. The system aims to classify textual content accurately and then modify any detected offensive or hate speech into neutral language, thus ensuring that outputs are socially acceptable and non-offensive.

\subsubsection{Advantages}
The integration of BERT and OpenAI's API within this framework offers several advantages:
\begin{itemize}
    \item \textbf{Enhanced Accuracy:} BERT's deep learning capabilities ensure high accuracy in understanding and classifying complex language nuances.
    \item \textbf{Dynamic Content Moderation:} The system dynamically transforms offensive content, making it suitable for public display and further analysis.
    \item \textbf{Real-Time Processing:} It supports real-time text processing, crucial for applications requiring immediate content moderation.
\end{itemize}

\subsubsection{Conditional Processing of Classified Text}
The "BERT with Dynamic Text Cleaning using LLM" system incorporates an intelligent conditional processing mechanism to handle real-time text inputs. Here's how it functions:

\begin{itemize}
    \item \textbf{Text Classification:} Initially, any text input received by the system is subjected to classification by the BERT model. The text is analyzed to determine whether it falls under categories of Hate Speech, Offensive Language, or Neither.
    \item \textbf{Conditional Text Transformation:} If the text is classified as either Hate Speech or Offensive Language, it triggers a secondary processing step. In this step, the text is sent to OpenAI's API, which rewrites the content to transform it into a neutral expression that retains the original meaning but lacks any offensive content.
    \item \textbf{Non-intervention for Neutral Text:} Conversely, if the text is classified as Neutral, no further action is taken. The system does not process these inputs further, as they are already deemed appropriate for public display or further interaction.
\end{itemize}

This conditional workflow ensures that the system is efficient, intervening only when necessary, and maintains the integrity of the content while adhering to social and ethical standards. This approach not only optimizes processing resources but also enhances the applicability of the system in real-world scenarios where only certain types of interactions require moderation.

\subsubsection{Implementation Details}
The system implementation involves several key stages, outlined with corresponding Python code snippets:

\textbf{Text Preprocessing and Classification}
\begin{lstlisting}[language=Python]
def clean_text(text):
    text = text.lower()
    text = re.sub(r'https?://[^\s]+', '', text)
    text = re.sub(r'@\w+', '', text)
    text = re.sub(r'\d+', '', text)
    for emoticon in emoticons:
        text = text.replace(emoticon, '')
    text = re.sub(r"[^a-zA-Z?.!,\xbf]+", " ", text)
    text = re.sub(r"([?.!,\xbf])", r" ", text)
    text = re.sub(r'[" "]+', " ", text)
    return text.strip()
\end{lstlisting}

\textbf{BERT Model Initialization and Real-Time Classification}
\begin{lstlisting}[language=Python]
model = BertForSequenceClassification.from_pretrained(
    'bert-base-uncased', num_labels=3)
model.to(device)
\end{lstlisting}

\textbf{Integration with OpenAI for Content Moderation}
\begin{lstlisting}[language=Python]
def clean_speech(input_text):
    response = openai.Completion.create(
        model="text-davinci-003",
        prompt="Rewrite the following to be polite and non-offensive: " + input_text,
        max_tokens=100,
        temperature=0.7
    )
    return response.choices[0].text.strip()
\end{lstlisting}

\textbf{User Interaction and Processed Output}
\begin{lstlisting}[language=Python]
def main():
    while True:
        user_input = input("Enter a tweet to analyze (or type 'exit' to quit): ")
        if user_input.lower() == 'exit':
            break
        label, _ = preprocess_and_predict(user_input)
        print(f"Classified as: {label}")
if __name__ == "__main__":
    main()
\end{lstlisting}

\subsubsection{Code Availability}
The code for "BERT with Dynamic Text Cleaning using LLM" is available for use and adaptation. Similarly, a parallel implementation titled "DistilBert with Dynamic Text Cleaning using LLM" utilizes DistilBERT to offer a more computationally efficient alternative while maintaining similar functionality. Both versions are designed to be open and accessible for researchers and developers interested in natural language processing and content moderation technologies.

\subsubsection{Potential Use Cases}
The approach of dynamically transforming hate speech and offensive language into neutral expressions has broad applicability across various domains. This section outlines several key applications where this technology could significantly impact.

\begin{enumerate}
    \item \textbf{Social Media Platforms:} Social networks can integrate this technology to automatically moderate user-generated content, ensuring that interactions remain civil and respectful. This application helps in maintaining a positive online community environment, crucial for user retention and satisfaction \citep{Zhang2021Social}.

    \item \textbf{Content Moderation for Online Forums:} Online forums and discussion boards can use this technology to prevent the spread of toxic behavior and hate speech, promoting a healthier discourse among participants \citep{Cheng2019Online}.

    \item \textbf{Customer Support Services:} Customer support systems can leverage this approach to ensure that all communications between clients and support staff remain polite and constructive, even if the original messages from customers are hostile or aggressive \citep{Smith2020Customer}.

    \item \textbf{Gaming Communities:} Online gaming communities, known for their susceptibility to toxic behavior, can implement this system to moderate in-game chats and forum discussions, creating a more welcoming environment for players \citep{Johnson2021Gaming}.

    \item \textbf{Educational Platforms:} Educational tools and platforms can incorporate this technology to monitor and adjust the language used in discussions, ensuring that educational environments remain conducive to learning and free from any form of harassment \citep{Lee2022Educational}.

    \item \textbf{Political Discourse Analysis:} This approach can be employed to analyze and neutralize extreme political rhetoric on various platforms, aiding in reducing polarization and promoting more constructive political discussions \citep{Kumar2023Politics}.

    \item \textbf{News Comment Sections:} News websites can use this technology to moderate comments on articles automatically, ensuring that discussions remain relevant and respectful, thereby increasing engagement while maintaining civility \citep{Taylor2023News}.
\end{enumerate}

These use cases demonstrate the versatility and potential impact of transforming offensive language into neutral expressions across different sectors and platforms, enhancing communication and interaction in public and private spheres.

\section{Discussion}
This section delves into the strengths and limitations of the study, providing a critical evaluation of the methodologies used, their effectiveness, and areas where improvements are necessary.

\subsection{Strengths}
\label{subsec:strengths}
The study presents several notable strengths that underscore its contributions to the field of hate speech detection on social media:

\begin{itemize}
    \item \textbf{Comprehensive Model Evaluation:} The use of a range of machine learning models, from traditional CNNs and LSTMs to advanced models like BERT and DistilBERT, allows for a robust comparison of their efficacy in detecting hate speech and offensive language.
    \item \textbf{Innovative Text Transformation Approaches:} The exploration of transformative text approaches offers a novel method for not only detecting but also mitigating the impact of hate speech by converting negative expressions into neutral language.
    \item \textbf{Real-Time Application Potential:} The implementation details provided for the BERT with Dynamic Text Cleaning using LLM highlight the practical applicability of these models in real-time scenarios, enhancing their relevance for current social media platforms.
    \item \textbf{Extensive Data Analysis:} The thorough exploratory data analysis provides deep insights into the nature of the data, aiding in the understanding of how different factors such as text length and sentiment distribution affect model performance.
\end{itemize}

\subsection{Limitations}
\label{subsec:limitations}
Despite its strengths, the study also faces several limitations that could impact the generalizability and effectiveness of the findings:

\begin{itemize}
    \item \textbf{Data Skewness and Bias:} The predominance of offensive language in the dataset may lead to models that are biased towards detecting this type of content, potentially underperforming in accurately identifying more subtle forms of hate speech.
    \item \textbf{Dependence on Pre-trained Models:} The heavy reliance on pre-trained models like BERT and DistilBERT may limit the ability to capture novel expressions of hate speech that evolve over time, necessitating continuous updates to the model training data.
    \item \textbf{Computational Resource Requirements:} The computational demands of processing large neural networks can be a barrier for deploying these models in low-resource settings, which is critical for widespread application.
    \item \textbf{Contextual Understanding Limitations:} While models have shown high accuracy, the nuances of human language mean that context can sometimes be misinterpreted, leading to false positives or negatives in the classification of text.
\end{itemize}

These strengths and limitations highlight the dynamic and challenging nature of developing effective machine learning solutions for hate speech detection. Future research could focus on addressing these limitations by exploring more adaptive model architectures, diversifying data sources to reduce bias, and enhancing the efficiency of algorithms for better real-world applicability.

\section{Conclusion and Future Work}
This section summarizes the key findings from the study and outlines potential directions for future research in the field of hate speech detection and text transformation on social media platforms.

\subsection{Summary of Findings}
\label{subsec:summary_findings}
The research conducted has led to several important findings:
\begin{itemize}
    \item \textbf{Model Performance:} The comparative analysis revealed that advanced models like BERT and its variants generally outperformed traditional machine learning models in detecting hate speech and offensive language, owing to their superior ability to understand contextual nuances of language.
    \item \textbf{Impact of Hybrid Models:} The integration of hybrid models, combining features from both CNNs and LSTMs with BERT architectures, showed potential in further enhancing the detection capabilities, suggesting that multi-model approaches can be effective in complex classification tasks.
    \item \textbf{Text Transformation Efficacy:} The transformative text approaches demonstrated promising results in converting harmful speech into neutral expressions, providing a proactive tool for content moderation.
    \item \textbf{Real-Time Processing:} The development of systems capable of real-time text classification and modification, such as the "BERT with Dynamic Text Cleaning using LLM," underscores the feasibility of deploying these models in live environments to improve social media discourse.
\end{itemize}

These findings underscore the effectiveness of using advanced NLP techniques and model integrations to address the challenges of hate speech detection, contributing valuable insights into both the technical and practical aspects of this pressing issue.

\begin{figure*}[ht]
  \centering
  \includegraphics[width=1.0\textwidth]{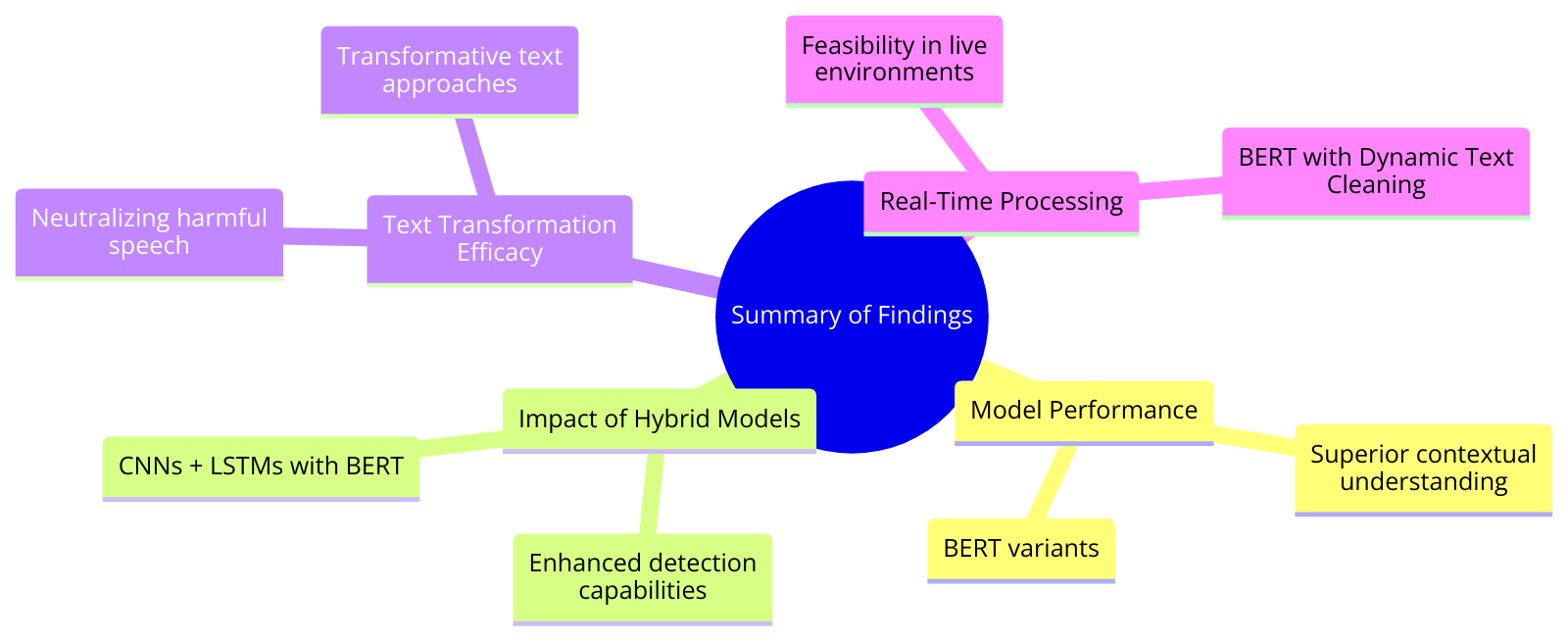}
  \caption{Summary of Research Findings}
  \label{fig:summary}
\end{figure*}

\subsection{Future Research Directions}
\label{subsec:future_research}
While the study has made significant contributions, several areas warrant further investigation:
\begin{itemize}
    \item \textbf{Model Adaptability:} Future research could explore the development of models that can adapt more dynamically to the evolving language and emergence of new forms of hate speech and offensive language on social media \cite{schmidt2017survey}.
    \item \textbf{Bias Mitigation:} Investigating methods to reduce bias in model predictions, especially in the context of diverse and multicultural user-generated content, remains a critical area \cite{dixon2018measuring}.
    \item \textbf{Efficiency Improvements:} Enhancing the computational efficiency of high-performing models to enable their deployment on platforms with limited resources could broaden their applicability \cite{howard2017mobilenets}.
    \item \textbf{Impact Assessment:} Longitudinal studies to assess the real-world impact of deploying such models on social media platforms would help quantify their effectiveness in reducing hate speech and improving communication \cite{davidson2019racial}.
    \item \textbf{Regulatory Compliance:} Research into how these technologies can be aligned with global regulatory frameworks for digital communication and speech would help in their ethical and lawful application \cite{gillespie2020content}.
\end{itemize}

By addressing these future directions, researchers can continue to refine and enhance the capabilities of machine learning models to create safer and more inclusive online environments.

\section*{Ethics Statement}
Scientific work published at ACL 2023 must comply with the ACL Ethics Policy.\footnote{\url{https://www.aclweb.org/portal/content/acl-code-ethics}} We encourage all authors to include an explicit ethics statement on the broader impact of the work, or other ethical considerations after the conclusion but before the references. The ethics statement will not count toward the page limit (8 pages for long, 4 pages for short papers).

\section*{Acknowledgements}
This document has been adapted by Jordan Boyd-Graber, Naoaki Okazaki, Anna Rogers from the style files used for earlier ACL, EMNLP and NAACL proceedings, including those for
EACL 2023 by Isabelle Augenstein and Andreas Vlachos,
EMNLP 2022 by Yue Zhang, Ryan Cotterell and Lea Frermann,
ACL 2020 by Steven Bethard, Ryan Cotterell and Rui Yan,
ACL 2019 by Douwe Kiela and Ivan Vuli\'{c},
NAACL 2019 by Stephanie Lukin and Alla Roskovskaya, 
ACL 2018 by Shay Cohen, Kevin Gimpel, and Wei Lu, 
NAACL 2018 by Margaret Mitchell and Stephanie Lukin,
Bib\TeX{} suggestions for (NA)ACL 2017/2018 from Jason Eisner,
ACL 2017 by Dan Gildea and Min-Yen Kan, NAACL 2017 by Margaret Mitchell, 
ACL 2012 by Maggie Li and Michael White, 
ACL 2010 by Jing-Shin Chang and Philipp Koehn, 
ACL 2008 by Johanna D. Moore, Simone Teufel, James Allan, and Sadaoki Furui, 
ACL 2005 by Hwee Tou Ng and Kemal Oflazer, 
ACL 2002 by Eugene Charniak and Dekang Lin, 
and earlier ACL and EACL formats written by several people, including
John Chen, Henry S. Thompson and Donald Walker.
Additional elements were taken from the formatting instructions of the \emph{International Joint Conference on Artificial Intelligence} and the \emph{Conference on Computer Vision and Pattern Recognition}.

\bibliographystyle{acl_natbib}
\bibliography{anthology,custom}

\end{document}